\documentclass[letterpaper]{article} 
\usepackage{aaai25}  
\usepackage{times}  
\usepackage{helvet}  
\usepackage{courier}  
\usepackage[hyphens]{url}  
\usepackage{graphicx} 
\usepackage{natbib}  
\usepackage{caption} 
\usepackage{subcaption}
\usepackage{dsfont}
\usepackage{makecell} %

\usepackage{algorithm}
\usepackage{algorithmic}

\usepackage{color}
\usepackage{booktabs}
\usepackage{multirow}
\usepackage{rotating}
\usepackage{amssymb}
\usepackage{amsmath, bm}

\usepackage{amsfonts}

\usepackage{newfloat}
\usepackage{listings}
\DeclareCaptionStyle{ruled}{labelfont=normalfont,labelsep=colon,strut=off} 
\floatstyle{ruled}
\newfloat{listing}{tb}{lst}{}
\floatname{listing}{Listing}
\title{RegMixMatch: Optimizing Mixup Utilization in Semi-Supervised Learning}

\author{
    Haorong Han\textsuperscript{\rm 1,\rm 2},
    Jidong Yuan\textsuperscript{\rm 1,\rm 2}\thanks{Corresponding Author},
    Chixuan Wei\textsuperscript{\rm 3},
    Zhongyang Yu\textsuperscript{\rm 1,\rm 2}
}
\affiliations{
    \textsuperscript{\rm 1}Key Laboratory of Big Data and Artificial Intelligence in Transportation, Ministry of Education, China\\
    \textsuperscript{\rm 2}School of Computer Science and Technology, Beijing Jiaotong University\\
    \textsuperscript{\rm 3}School of Data Science and Intelligent Media,
 Communication University of China\\
    \{harohan, yuanjd, zhongyangyu\}@bjtu.edu.cn,
    chxwei@cuc.edu.cn

}

\usepackage{bibentry}

\begin{document}

\maketitle

\begin{abstract}

Consistency regularization and pseudo-labeling have significantly advanced semi-supervised learning (SSL). Prior works have effectively employed Mixup for consistency regularization in SSL. However, our findings indicate that applying Mixup for consistency regularization may degrade SSL performance by compromising the purity of artificial labels. Moreover, most pseudo-labeling based methods utilize thresholding strategy to exclude low-confidence data, aiming to mitigate confirmation bias; however, this approach limits the utility of unlabeled samples. To address these challenges, we propose RegMixMatch, a novel framework that optimizes the use of Mixup with both high- and low-confidence samples in SSL. First, we introduce semi-supervised RegMixup, which effectively addresses reduced artificial labels purity by using both mixed samples and clean samples for training. Second, we develop a class-aware Mixup technique that integrates information from the top-2 predicted classes into low-confidence samples and their artificial labels, reducing the confirmation bias associated with these samples and enhancing their effective utilization. Experimental results demonstrate that RegMixMatch achieves state-of-the-art performance across various SSL benchmarks.
\end{abstract}

\begin{links}
\link{Code}{https://github.com/hhrd9/regmixmatch}
\end{links}

\section{Introduction}
\label{Introduction}

Semi-supervised learning (SSL) aims to leverage a small amount of labeled data to enable the model to extract useful information from a large volume of unlabeled data during training. FixMatch \cite{sohn2020fixmatch} and its variants \cite{zhang2021flexmatch,xu2021dash,wang2023freematch,huang2023flatmatch} have demonstrated competitive results in SSL by retaining pseudo-labels \cite{lee2013pseudo} for high-confidence samples using a threshold strategy combined with consistency regularization. However, these methods heavily rely on data augmentation techniques for consistency regularization, and their pseudo-labeling approach is constrained to high-confidence samples alone. This raises two critical questions: 1) \textit{Are existing data augmentation techniques optimal for SSL?} 2) \textit{Are the low-confidence samples filtered by the threshold truly unusable?} At the core of these questions is the challenge of effectively utilizing both high- and low-confidence samples. In this paper, we aim to address these questions by proposing solutions that optimize the use of the Mixup technique \cite{zhang2018mixup} in SSL.

\begin{figure}
  \centering
  \captionsetup[subfigure]{font=scriptsize,labelfont=scriptsize}
  \begin{subfigure}{0.49\linewidth}
    \includegraphics[width=\linewidth]{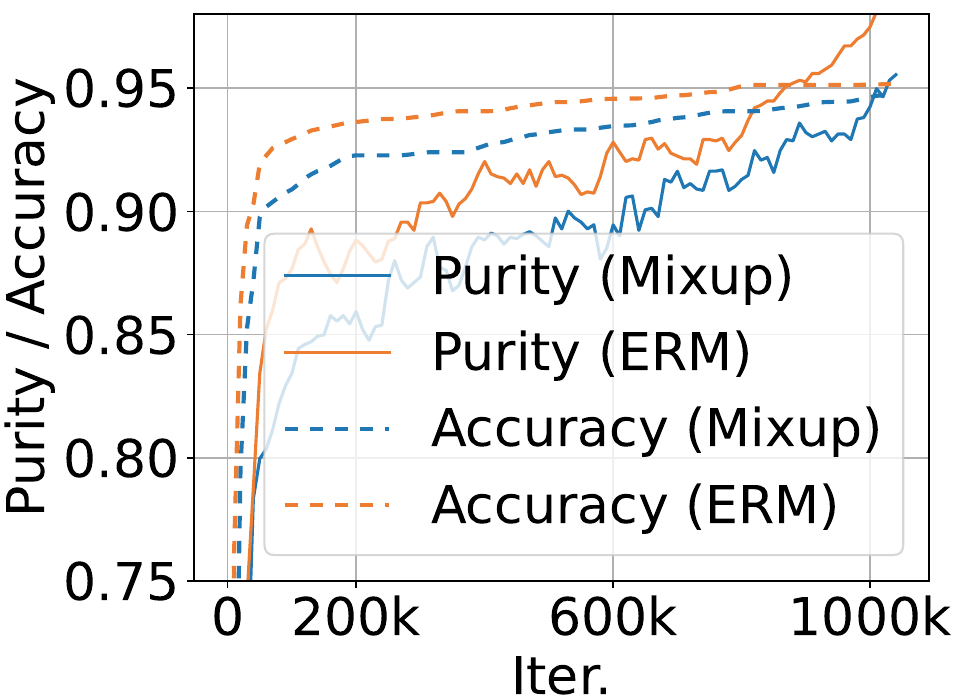}
    \caption{Purity and accuracy.} \label{fig:confi}
  \end{subfigure}%
  \hfill
  \begin{subfigure}{0.49\linewidth}
    \includegraphics[width=\linewidth]{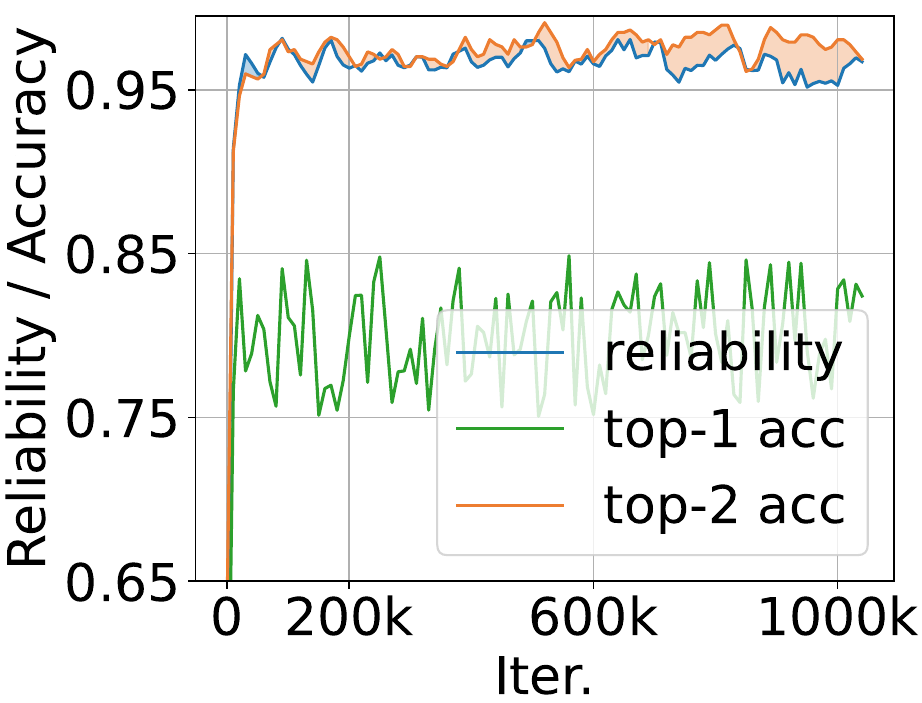}
    \caption{Reliability and top-\textit{n} accuracy.} \label{fig:topn}
  \end{subfigure}

  \caption{ Motivating example of RegMixMatch. (a) Mixup's confidence-reducing behavior undermines SSL. (b) The top-2 accuracy rate across all unlabeled data surpasses the reliability of pseudo labels.}\label{fig:st_point}
\end{figure}

Mixup, which generates augmented samples by interpolating between randomly shuffled data, has been introduced into SSL \cite{berthelot2019mixmatch, berthelot2019remixmatch} to enhance model generalization, yielding significant results. However, recent studies \cite{wen2021combining, pinto2022using} have revealed that Mixup can induce confidence-reducing behavior because it inherently predicts interpolated labels for every input. We argue that this confidence reduction is problematic in SSL because creating effective artificial labels for unlabeled samples is crucial for training. Methods like sharpening \cite{berthelot2019remixmatch} and pseudo-labeling \cite{sohn2020fixmatch}, which harden softmax prediction probabilities, have been shown to be effective. This suggests that \textit{SSL expects to achieve purer artificial labels through low-entropy prediction probabilities, thereby better supporting model training.} However, Mixup increases the entropy of prediction probabilities, adding noise to artificial labels and consequently affecting SSL classification performance due to the lack of high-purity labels. As illustrated in Figure \ref{fig:st_point}(a), the purity (the proportion of artificial labels where the highest predicted probability exceeds a threshold, reflecting the model's prediction confidence) and test accuracy are compared when FixMatch is trained using Empirical Risk Minimization (ERM) \cite{vapnik1991principles} and Mixup methods with pseudo labels. The results indicate that Mixup reduces prediction confidence, decreasing the number of high-purity artificial labels and leading to lower classification performance than ERM, contrasting its success in supervised learning \cite{zhang2018mixup}. RegMixup \cite{pinto2022using} addresses this by using Mixup as a regularizer, training with both unmixed and mixed samples to mitigate the high-entropy issue. Inspired by this, and to counter the reduction in artificial label purity caused by Mixup's high-entropy behavior in SSL, we propose Semi-supervised RegMixup (SRM). SRM combines SSL’s unique pseudo-labeling technique with weak-to-strong consistency regularization, applying RegMixup within SSL. However, pseudo-labeling alone only allows SRM to effectively utilize high-confidence samples. In the following section, we will demonstrate how to effectively utilize low-confidence data.

Pseudo-labeling typically employs a fixed or dynamic threshold \cite{zhang2021flexmatch, xu2021dash, wang2023freematch, wei2023time} to filter out low-confidence samples, retaining pseudo labels only for those with confidence above the threshold. While this strategy fully exploits high-confidence samples, it leaves samples with confidence below the threshold underutilized, as the supervisory information from these samples is generally noisier. However, unlike previous approaches that focus exclusively on leveraging high-confidence samples, we argue that low-confidence samples also hold significant potential. In Figure \ref{fig:st_point}(b), we present the reliability of pseudo labels (i.e., the accuracy of predictions for unlabeled data that surpass the threshold) during FixMatch training, alongside the top-1 and top-2 accuracy rates of the model's predictions for all unlabeled data. Notably, while the top-1 accuracy is relatively low, the top-2 accuracy exceeds the reliability of pseudo labels. This indicates that by developing methods that incorporate information from the top-2 classes into the samples and their artificial labels, we could enhance the utilization of a broader range of samples, thereby generating more reliable artificial labels even for those with low initial confidence.

To this end, we introduce a heuristic method called Class-Aware Mixup (CAM), designed to effectively utilize low-confidence samples by generating mixed samples for model training. Since artificial labels for low-confidence samples often contain significant noise due to uncertain predictions, mixing these samples with high-confidence samples of the same predicted class helps to reduce noise in the artificial labels, thereby improving their quality for these challenging samples. Our approach, dubbed \textbf{RegMixMatch}, integrates SRM and CAM, both of which leverage Mixup to exploit high- and low-confidence samples, respectively. To validate the effectiveness of this approach, we compare the training time and classification performance of RegMixMatch with previous Mixup-based SSL methods in our experiments, demonstrating significant improvements. The key contributions of our work can be summarized as follows:
\begin{itemize}
\item We discover and verify that Mixup reduces the purity of artificial labels in SSL, ultimately leading to a decline in model performance.
\item We propose SRM, which applies RegMixup to SSL by combining pseudo-labeling and consistency regularization to address the aforementioned issue.
\item We introduce CAM for low-confidence samples, aimed at mitigating confirmation bias and fully exploiting the potential of low-confidence data.
\item We conduct extensive experiments to validate the performance of RegMixMatch, achieving state-of-the-art results in most scenarios.
\end{itemize}

\section{Related Work}

Since the advent of deep learning, SSL has experienced rapid advancement. In this paper, we present an overview of SSL from two key perspectives: consistency regularization and pseudo-labeling. Additionally, we provide a brief outline of data augmentation techniques based on mixing strategies.

\paragraph{Consistency Regularization.} 
Consistency regularization \cite{bachman2014learning} strengthens SSL by enforcing the principle that a classifier should maintain consistent class predictions for an unlabeled example, even after the application of data augmentation. This principle has led to the development of various SSL algorithms based on different augmentation techniques. Traditional SSL methods typically employ softmax or sharpened softmax outputs to supervise perturbed data, thereby achieving consistency regularization. For instance, Temporal Ensembling \cite{laine2017temporal} ensures model consistency by periodically updating the output mean and minimizing the difference in predictions for the same input across different training epochs. Mean Teacher \cite{tarvainen2017mean} enhances learning targets by generating a superior teacher model through the exponential moving average of model weights. VAT \cite{miyato2018virtual} introduces adversarial perturbations to input data, increasing the model's robustness. More recently, MixMatch \cite{berthelot2019mixmatch} and ReMixMatch \cite{berthelot2019remixmatch} have incorporated the Mixup technique to create augmented samples, thereby boosting generalization performance. ReMixMatch, along with UDA \cite{xie2020unsupervised}, also integrates advanced image augmentation techniques such as AutoAugment \cite{cubuk2019autoaugment} and RandAugment \cite{cubuk2020randaugment}, which generate heavily distorted yet semantically intact images, further enhancing classification performance.

\paragraph{Pseudo-Labeling.}
Pseudo-labeling \cite{lee2013pseudo} has been demonstrated to offer superior supervision for augmented samples in SSL. To address the issue of confirmation bias inherent in pseudo-labeling, confidence-based thresholding techniques have been developed to enhance the reliability of pseudo labels. FixMatch \cite{sohn2020fixmatch} employs a fixed threshold to generate pseudo labels from high-confidence predictions on weakly augmented images, ensuring consistency with their strongly augmented counterparts. FlexMatch \cite{zhang2021flexmatch} and FreeMatch \cite{wang2023freematch} introduce class-specific thresholds that adapt based on each class's learning progress. Dash \cite{xu2021dash} dynamically adjusts the threshold during training to refine the filtering of pseudo labels. MPL \cite{pham2021meta} leverages a teacher network to produce adaptive pseudo labels for guiding a student network on unlabeled data. SoftMatch \cite{chen2023softmatch} tackles the trade-off between pseudo-label quality and quantity through a weighting function. Building on pseudo-labeling, FlatMatch \cite{huang2023flatmatch} minimizes a cross-sharpness measure to ensure consistent learning across labeled and unlabeled datasets, while SequenceMatch \cite{nguyen2024sequencematch} introduces consistency constraints between augmented sample pairs, reducing discrepancies in the model’s predicted distributions across different augmented views. Although pseudo-labeling enhances supervision and overall performance in these methods, its inherent limitations restrict the utilization to primarily high-confidence samples.

In this paper, we employ the SRM, specifically designed for SSL, to achieve consistency regularization by addressing the issue of reduced artificial label purity through the retention of a portion of clean samples for training. Additionally, we explore methods to utilize low-confidence samples through CAM. The differences from previous Mixup-based approaches are detailed in the appendix.

\paragraph{Mixing-Based Data Augmentation.} Mixup \cite{zhang2018mixup} generates augmented data by performing a linear interpolation between samples and their corresponding labels, aiming to achieve smoother decision boundaries. Unlike pixel-level Mixup, Manifold Mixup \cite{verma2019manifold} applies interpolation within the model’s hidden layers, allowing regularization at different representation levels. CutMix \cite{yun2019cutmix} combines the principles of Mixup and Cutout \cite{devries2017improved} by cutting a region from one image and pasting it onto another, thereby mixing samples while preserving the local structure of the images. SaliencyMix \cite{uddin2021saliencymix} further improves upon CutMix by detecting salient regions, ensuring that the cropped regions express label-relevant features, thus enhancing performance. ResizeMix \cite{qin2020resizemix} simplifies the process by resizing the source image to a small patch and pasting it onto another image, effectively retaining essential image information. For a more detailed comparison of different image mixing methods, we recommend reviewing OpenMixup \cite{li2023openmixup}. In this work, ResizeMix is chosen as the primary mixing method, with results based on Mixup presented in the appendix for comparison.

\section{Preliminary: From Mixup To RegMixup}
\label{section2}

Given a training dataset \( \mathcal{D} = \{(x_i, y_i)\}_{i=1}^n \), ERM \cite{vapnik1991principles} uses the empirical data distribution \( P_\delta(x,y) \) from the training set to approximate the true data distribution \( P(x,y) \):
\begin{equation}
P_\delta(x,y) = \frac{1}{n} \sum_{i=1}^n \delta(x=x_i, y=y_i),
\end{equation}
where \( \delta(x=x_i, y=y_i) \) denotes the Dirac mass centered at \( (x_i, y_i) \). The loss function is then minimized over the data distribution \( P_\delta \). In contrast to ERM, Vicinal Risk Minimization (VRM) \cite{chapelle2000vicinal} employs strategies to generate new data distributions in the vicinity of each sample, thereby estimating a richer distribution that provides a more informed risk assessment within the neighborhood around each sample. Based on VRM, Mixup \cite{{zhang2018mixup}} achieves a convex combination of samples to approximate the true data distribution, enhancing the model's generalization performance. The vicinal distribution of Mixup can be represented as
\begin{equation}
P_v(x,y) = \frac{1}{n} \sum_{i=1}^n \delta(x=\bar{x}_i, y=\bar{y}_i),
\end{equation}
where \( \bar{x}_i = \lambda x_i + (1-\lambda)x_j \) and \( \bar{y}_i = \lambda y_i + (1-\lambda)y_j \). For each \( (x_i, y_i) \) feature-target vector, the corresponding \( (x_j, y_j) \) is drawn at random from the training data. \( \lambda \) follows a Beta distribution with parameters \( (\alpha, \alpha) \). The larger the \( \alpha \), the closer \( \lambda \) is to 0.5, resulting in stronger interpolation.

Research on out-of-distribution (OOD) detection by RegMixup \cite{pinto2022using} reveals that Mixup can induce confidence-reducing behavior, impairing the model's ability to effectively differentiate between in-distribution and out-of-distribution samples. To counteract this issue, they propose that by using Mixup to generate mixed samples while retaining clean samples for training, the high-entropy behavior can be mitigated, thereby enhancing the model's robustness to OOD samples. In this context, the approximate data distribution \( P_v(x,y) \) can be constructed as
\begin{footnotesize}
\begin{equation}
P_v(x,y) = \frac{1}{n} \sum_{i=1}^n \left[\delta(x=x_i, y=y_i) + \delta(x=\bar{x}_i, y=\bar{y}_i)\right].
\end{equation}
\end{footnotesize}Based on this distribution, the following loss function is minimized:
\begin{equation}
H(y_i, p_\theta(y \mid x_i)) + H(\bar{y}_i, p_\theta(y \mid \bar{x}_i)),
\label{eq:regmixup}
\end{equation}
where \( H(\cdot,\cdot) \) denotes the standard cross-entropy loss, and \( p_\theta \) represents the softmax output of a neural network parameterized by \( \theta \). Later, we will introduce the proposed SRM, which integrates RegMixup into SSL by utilizing pseudo-labeling to address the reduction in artificial label purity caused by Mixup.

\section{RegMixMatch}
\label{section3}

\begin{figure*}[t]
    \centering
    \includegraphics[width=0.99\textwidth]{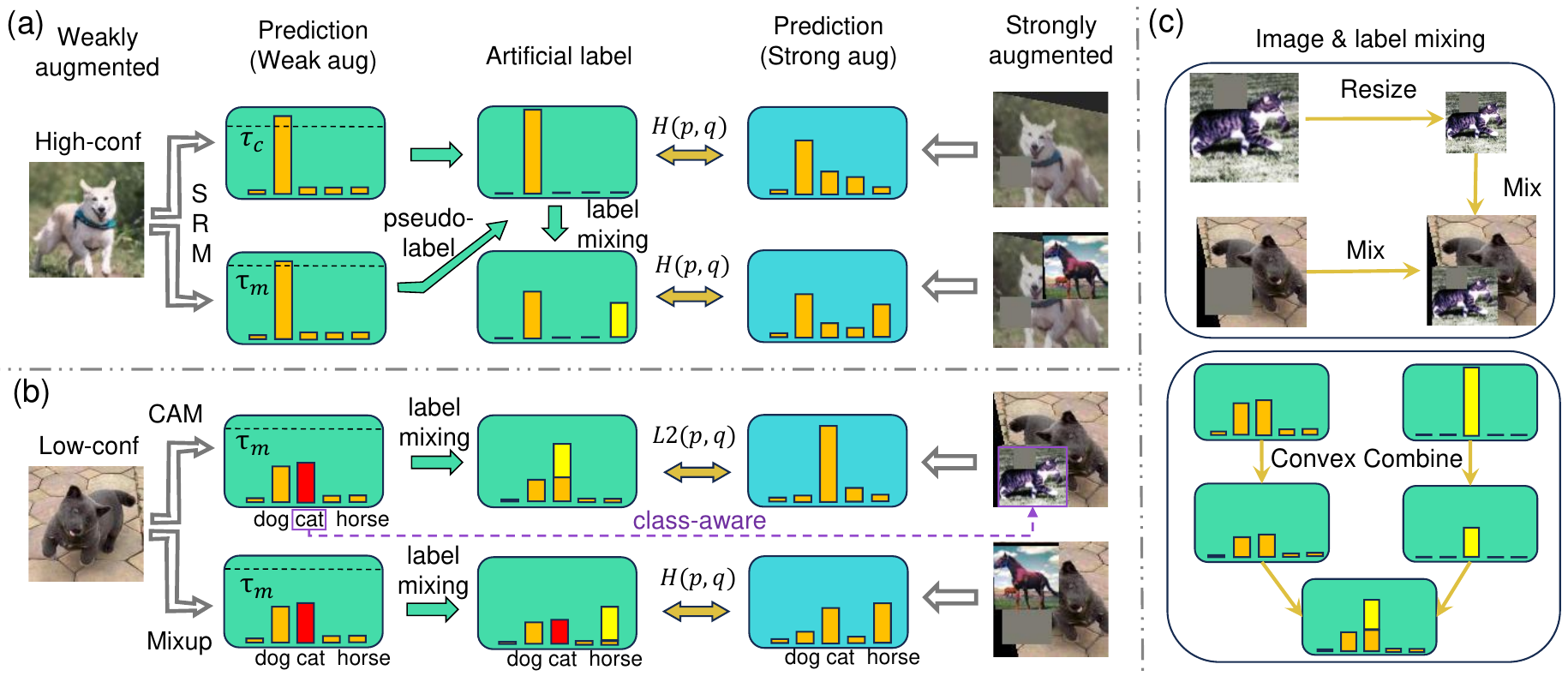}
    \caption{Overview of RegMixMatch. (a) shows the main idea of SRM. A weakly augmented image is fed into the model to obtain predictions. If the prediction confidence surpasses the threshold $\tau_c$, a pseudo label is used to compute consistency loss against the model's prediction on the corresponding strongly augmented view. If the confidence also exceeds the threshold $\tau_m$, its strongly augmented view and pseudo label are mixed with those of another high-confidence sample to implement Mixup. (b) illustrates why CAM works. When a low-confidence sample is mispredicted, CAM converts noise (red part) from artificial label into useful information, while MixUp suffers from it. (c) demonstrates how ResizeMix generates mixed images and labels.}
    \label{fig:box}
\end{figure*}

Given a batch of labeled data containing \( B \) samples \( \mathcal{X} = \{(x_b^l , y_b) : b \in (1, \ldots, B)\} \) and a batch of unlabeled data containing \( \mu B \) samples \( \mathcal{U} = \{(x_b^u) : b \in (1, \ldots, \mu B)\} \),  \( \mu \) is a hyperparameter that determines the relative sizes of \( \mathcal{X} \) and \( \mathcal{U} \). The supervised loss for the labeled samples in SSL methods is typically constructed as the standard cross-entropy loss between the model's predictions and the true labels:
\begin{equation}
\mathcal{L}_s = \frac{1}{B} \sum_{b=1}^{B} H(y_b, p_m(y|x_b^l)),
\label{eq:ls}
\end{equation}
where \( p_m(y|x) \) is the predicted class distribution produced by the model for input \( x \).

Despite the variety of data augmentation methods, they all follow consistency regularization when utilizing unlabeled data: the expectation is that the model outputs similar results for differently perturbed versions of the unlabeled data. Based on this idea, the consistency loss is typically constructed as
\begin{equation}
\frac{1}{\mu B} \sum_{b=1}^{\mu B} \ell_c(p_m(y|\alpha(x_b^u)), p_m(y|\alpha(x_b^u))),
\label{eq:closs}
\end{equation}
where \( \ell_c \) is the consistency loss, which can be cross-entropy or \( \ell \)-2  loss. \( \alpha(x) \) represents a random augmentation of input \( x \), meaning the two terms in Equation \ref{eq:closs} can have different values due to the different random augmentations applied.

\subsection{From RegMixup to Semi-supervised RegMixup}
The overall architecture of SRM is depicted in Figure \ref{fig:box}(a). In supervised learning, RegMixup employs cross-entropy loss on both clean and mixed samples. However, in SSL, a substantial portion of the data remains unlabeled. By leveraging pseudo-labeling, we retain pseudo labels for unlabeled samples that exceed a specific confidence threshold. In this section, we will illustrate how the two terms in Equation \ref{eq:regmixup} are adapted for SSL.

For an unlabeled image \( x_b^u \), we first apply a weak augmentation \( \alpha(x_b^u) \) and compute the model's predicted class distribution \( q_b = p_m(y|\alpha(x_b^u)) \). The class with the highest probability, denoted as \( \hat{q}_b = \arg\max(q_b) \), is then assigned as the pseudo label for the corresponding strongly augmented view \( \mathcal{A}(x_b^u)\). Consequently, we transform the RegMixup loss function for clean samples into the following form:
\begin{footnotesize}
\begin{equation}
\mathcal{L}_u = \frac{1}{\mu B} \sum_{b=1}^{\mu B} \mathds{1}(\max{(q_b)} \geq \tau_c) H(\hat{q}_b, p_m(y|\mathcal{A}(x_b^u))),
\label{eq:lu}
\end{equation}
\end{footnotesize}where \( \tau_c \) is the threshold above which a pseudo-label is retained, and \( \max{(q_b)} \) represent the maximum predicted probability of the weak augmentation \( \alpha(x_b^u) \).

To implement Mixup, we also apply thresholding to retain pseudo labels only for high-confidence unlabeled images. Specifically, for a weakly augmented view \( \alpha(x_i^u) \) of an unlabeled image within a data batch, we retain the pseudo label \( \hat{q}_i \) only if its confidence exceeds the threshold \( \tau_m \). High-confidence images from this batch are then grouped into a set denoted as $ \mathcal{H} = \{x_b^u|\max(q_b) > \tau_m\}$. For the loss of mixed data, as shown in Figure \ref{fig:box}(a), given a strongly augmented view \( \mathcal{A}(x_i^u) \) and its pseudo label \( \hat{q}_i \) from the set \( \mathcal{H} \), we randomly select \( \mathcal{A}(x_j^u) \) and \( \hat{q}_j \) from the same set \( \mathcal{H} \), generating mixed images and labels using ResizeMix technique for the calculation of \( H(\cdot,\cdot) \). Thus, we transform the loss function of RegMixup for mixed samples into
\begin{equation}
\mathcal{L}_m = \frac{1}{|\mathcal{H}|} \sum_{i=1}^{|\mathcal{H}|} H(\hat{q}_i \oplus \hat{q}_j, p_m(y|\mathcal{A}(x_i^u) \oplus \mathcal{A}(x_j^u))),
\label{eq:lm}
\end{equation}
where the symbol \( \oplus \) denotes the image or label mixing operation, and \( |\cdot| \) represents the number of elements in the set. 

Note that SRM employs two different thresholds, \( \tau_c \) and \( \tau_m \). It has been observed that when \( \tau_m \) is close to or lower than \( \tau_c \), the model is prone to significant confirmation bias, particularly during the early training stages. To mitigate this, \( \tau_m \) is set to a higher value. Additionally, as described in the preliminary, \( \alpha \) is used to control the mixing intensity; a higher \( \alpha \) results in a stronger mixing intensity. Our parameter \( \alpha_h \), which controls the mixing intensity of high-confidence samples, can be set to a higher value than previous Mixup-based SSL methods due to the retention of clean samples, allowing for better generalization performance. 

In summary, we select high-confidence samples as clean samples using a high threshold \( \tau_c \) and retain their pseudo labels to compute the consistency loss. Concurrently, we set an even higher threshold \( \tau_m \) to filter and retain the high-confidence samples and their pseudo labels for implementing Mixup. Consequently, we can transform Equation \ref{eq:regmixup} into the following form to achieve the transition from RegMixup to SRM:
\begin{equation}
\mathcal{L}_{rm} = \mathcal{L}_u + \mathcal{L}_m.
\end{equation}

\subsection{Class-Aware Mixup}
\label{sec:cam}
Pseudo-labeling enables our SRM to effectively leverage high-confidence samples. However, low-confidence samples are often discarded due to the uncertainty in their predictions. Our proposed CAM method facilitates the utilization of these low-confidence samples while maintaining the quality of the artificial labels.

Figure \ref{fig:box}(b) provides a conceptual overview of our CAM. Mixup randomly selects a target image, which can belong to any class, to blend with the source image. For low-confidence unlabeled samples, where the predicted class often diverges from the true class, randomly selecting a target image can lead to confirmation bias. This is because the mixed label may incorporate noisy features from the predicted class, while the mixed image does not reflect these features. However, by limiting the samples mixed with low-confidence ones to high-confidence samples sharing the same predicted classes (i.e., class-aware), the noise can be reduced due to the inclusion of predicted class features. Additionally, Figure \ref{fig:st_point}(b) shows that the top-2 accuracy is high, indicating a  strong likelihood that the correct class is among the top-2 predicted classes in the mixed labels, making the inclusion of predicted class information sufficient. 

We denote the complement of \( \mathcal{H} \) as \( \mathcal{H}^c \). Given a strongly augmented view \( \mathcal{A}(x_i^u) \) of a low-confidence data in \( \mathcal{H}^c \) and its softmax output \( q_i \), we use the strongly augmented view \( \mathcal{A}(x_j^u) \) and its pseudo label \( \hat{q}_j \) of the class-aware sample ($ \hat{q}_j = \arg\max(q_i) $) from the high-confidence set \( \mathcal{H} \) , generating mixed images and labels using ResizeMix technique for the calculation of  \( \ell \)-2 loss (as shown in Equation \ref{eq:lcm}).

\begin{footnotesize}
\begin{equation}
\mathcal{L}_{cm} = \frac{1}{|\mathcal{H}^c|} \sum_{i=1}^{|\mathcal{H}^c|} \left\| q_i \oplus \hat{q}_j - p_m(y|\mathcal{A}(x_i^u) \oplus \mathcal{A}(x_j^u)) \right\|_2^2.
\label{eq:lcm}
\end{equation}
\end{footnotesize}To reduce the negative impact of uncertain predictions, we use softmax outputs as artificial labels for low-confidence data and apply \( \ell \)-2 loss instead of cross-entropy loss. Importantly, the parameter \( \alpha_l \) in CAM, which controls the mixing intensity for low-confidence samples, is set higher than \( \alpha_h \). This choice helps to minimize the risk of the model overfitting to specific noisy samples, a point that will be further explored in the sensitivity analysis. Although the top-$n$ ($n \geq 3$) accuracy of the model's predictions would be higher, our experiments show that mixing more than two samples does not yield additional improvements. Ultimately, CAM enables the incorporation of information from the top-2 classes into the samples and their artificial labels, thereby facilitating the utilization of low-confidence samples. 

\begin{algorithm}[t]
\caption{\textbf{RegMixMatch Algorithm}}
\label{alg:RegMixMatch}
\begin{algorithmic}[1]
\STATE \textbf{Input:} Labeled batch $\mathcal{X} = \{(x_b^l, y_b) : b \in (1, \dots, B)\}$, 
unlabeled batch $\mathcal{U} = \{x_b^u : b \in (1, \dots, \mu B)\}$, 
confidence threshold $\tau_c$ and $\tau_m$, 
mixing intensities $\alpha_h$ and $\alpha_l$ for high- and low-confidence data
\STATE Calculate $\mathcal{L}_s$ using Equation \ref{eq:ls} \textcolor{black}{\COMMENT{Loss for labeled data}}
\FOR{$b = 1$ to $\mu B$}
    \STATE $q_b = p_m(y|\alpha(x_b^u); \theta)$ \textcolor{black}{\COMMENT{Prediction after weak augmentation}}
    \STATE $\hat{q}_b = \arg\max(q_b)$ \textcolor{black}{\COMMENT{Pseudo labels for unlabeled data}}
\ENDFOR
\STATE Calculate $\mathcal{L}_u$ using Equation \ref{eq:lu} \textcolor{black}{\COMMENT{Loss for clean unlabeled data}}
\STATE Construct high-confidence set $ \mathcal{H} = \{x_b^u|\max(q_b) > \tau_m\}$ 
\STATE Construct low-confidence set $ \mathcal{H}^c = \mathcal{U} \setminus \mathcal{H}$ 
\STATE Calculate $\mathcal{L}_m$ using Equation \ref{eq:lm} \textcolor{black}{\COMMENT{Loss for mixed high-confidence unlabeled data}}
\STATE $\mathcal{L}_{rm} = \mathcal{L}_m + \mathcal{L}_u$ \textcolor{black}{\COMMENT{Using Mixup as a regularizer}}
\STATE Calculate $\mathcal{L}_{cm}$ using Equation \ref{eq:lcm} \textcolor{black}{\COMMENT{Loss for mixed low-confidence unlabeled data}}
\RETURN $\mathcal{L}_s + \mathcal{L}_{rm} + \mathcal{L}_{cm}$
\end{algorithmic}
\end{algorithm}

Overall, our proposed SRM employs pseudo-labeling and consistency regularization for high-confidence data, applying Mixup to generate mixed samples while preserving unmixed samples for training in SSL. SRM enhances model generalization by leveraging Mixup, while simultaneously addressing the issue of degraded artificial label purity due to Mixup's high-entropy behavior. For low-confidence samples, our CAM method mixes them with samples from specific classes, effectively utilizing these challenging samples while mitigating confirmation bias.  The full algorithm of RegMixMatch is shown in Algorithm \ref{alg:RegMixMatch}. The overall loss of RegMixMatch is formulated as

\begin{equation}
\mathcal{L} = \mathcal{L}_s + \mathcal{L}_{rm} + \mathcal{L}_{cm}.
\end{equation}

\begin{table*}[t]
	\centering
	\fontsize{7pt}{6.5pt}\selectfont
	\begin{tabular}{l|cccc|ccc|ccc|cc}
		\toprule
		Dataset & \multicolumn{4}{c|}{CIFAR10}& \multicolumn{3}{c|}{CIFAR100}& \multicolumn{3}{c|}{SVHN} & \multicolumn{2}{c}{STL10} \\ \cmidrule(r){1-1}\cmidrule(lr){2-5}\cmidrule(lr){6-8}\cmidrule(lr){9-11}\cmidrule(l){12-13}
		
		\# Label & \multicolumn{1}{c}{10} & \multicolumn{1}{c}{40} & \multicolumn{1}{c}{250}  & \multicolumn{1}{c|}{4000} & \multicolumn{1}{c}{400}  & \multicolumn{1}{c}{2500}  & \multicolumn{1}{c|}{10000} & \multicolumn{1}{c}{40}  & \multicolumn{1}{c}{250}   & \multicolumn{1}{c|}{1000} & \multicolumn{1}{c}{40}  & \multicolumn{1}{c}{1000}\\ \cmidrule(r){1-1}\cmidrule(lr){2-5}\cmidrule(lr){6-8}\cmidrule(lr){9-11}\cmidrule(l){12-13}
		
		VAT \cite{miyato2018virtual} & 79.81 & 74.66 & 41.03 & 10.51 & 85.20 & 46.84 & 32.14 & 74.75 & 4.33 & 4.11 & 74.74  & 37.95 \\
		Mean Teacher \cite{tarvainen2017mean} & 76.37 & 70.09 & 37.46 & 8.10 & 81.11 & 45.17 & 31.75 & 36.09 & 3.45 & 3.27 & 71.72 & 33.90 \\
		MixMatch \cite{berthelot2019mixmatch} & 65.76 & 36.19 & 13.63 & 6.66 & 67.59 & 39.76 & 27.78 & 30.60 & 4.56 & 3.69 & 54.93 & 21.70 \\
		ReMixMatch \cite{berthelot2019remixmatch} & 20.77 & 9.88 & 6.30 & 4.84 & 42.75 & 26.03 &  20.02 & 24.04 & 6.36 & 5.16 & 32.12 & 6.74 \\
		UDA \cite{xie2020unsupervised} & 34.53 & 10.62 & 5.16 & 4.29 & 46.39 & 27.73 & 22.49 & 5.12 & 1.92 & 1.89 & 37.42 & 6.64 \\
		FixMatch \cite{sohn2020fixmatch} & 24.79 & 7.47 & 4.86 & 4.21 & 46.42 & 28.03 & 22.20 & 3.81 & 2.02 & 1.96 & 35.97 & 6.25 \\
		Dash \cite{xu2021dash} & 27.28 & 8.93 & 5.16 & 4.36 & 44.82 & 27.15 & 21.88 & 2.19 & 2.04 & 1.97 & 34.52 & 6.39 \\
		MPL \cite{pham2021meta} & 23.55 & 6.62 & 5.76 & 4.55 & 46.26 & {27.71} & 21.74 & 9.33 & 2.29 & 2.28 & 35.76 & 6.66 \\
		FlexMatch \cite{zhang2021flexmatch} & 13.85 & 4.97 & 4.98 & 4.19 & 39.94 & 26.49 & 21.90 & 8.19 & 6.59 & 6.72 & 29.15 & 5.77 \\
		SoftMatch \cite{chen2023softmatch} & - & 4.91 & 4.82 & 4.04 & \underline{37.10} & 26.66 & 22.03 & 2.33 & - & 2.01 & 21.42 & 5.73 \\
		FreeMatch \cite{wang2023freematch} & \underline{8.07} & 4.90 & 4.88 & 4.10 & 37.98 & 26.47 & 21.68 & 1.97 & 1.97 & 1.96 & 15.56 & 5.63 \\
		SequenceMatch \cite{nguyen2024sequencematch} & - & \underline{4.80} & 4.75 & 4.15 & 37.86 & 25.99 & 20.10 & \underline{1.96} & 1.89 & 1.79 & \underline{15.45} & 5.56 \\                
		FlatMatch \cite{huang2023flatmatch} & 15.23 & 5.58 & \underline{4.22} & \underline{3.61} & 38.76 & \underline{25.38} & \textbf{19.01} & 2.46 & \textbf{1.43} & \textbf{1.41} & 16.20 & \underline{4.82} \\
		RegMixMatch & \textbf{4.35} & \textbf{4.24} & \textbf{4.21} & \textbf{3.38} & \textbf{35.27} & \textbf{23.78} & \underline{19.41} & \textbf{1.81} & \underline{1.77} & \underline{1.79} & \textbf{11.74} & \textbf{4.66} \\
		\midrule
		Fully-Supervised    & \multicolumn{4}{c|}{4.62} & \multicolumn{3}{c|}{19.30}  & \multicolumn{3}{c|}{2.13} & \multicolumn{2}{c}{-}\\
		\bottomrule
	\end{tabular}
	\caption{\small Error rates on CIFAR10/100, SVHN, and STL10 datasets. The fully-supervised results of STL10 are unavailable since we do not have label information for its unlabeled data. The best results are highlighted with \textbf{Bold} and the second-best results are highlighted with \underline{underline}. We ran each task three times, and the results with standard deviation are presented in the appendix.}
	\label{tab:comparsion}
\end{table*}

\section{Experiments}
\label{section4}

In this section, we present an extensive experimental evaluation of the proposed RegMixMatch method. We assess its performance across a variety of widely-used SSL datasets, including CIFAR-10/100 \cite{krizhevsky2009learning}, SVHN \cite{netzer2011reading}, STL-10 \cite{coates2011analysis}, and ImageNet \cite{deng2009imagenet}, under different labeled data conditions. The experimental results are benchmarked against 13 established SSL algorithms, including VAT \cite{miyato2018virtual}, Mean Teacher \cite{tarvainen2017mean}, MixMatch \cite{berthelot2019mixmatch}, ReMixMatch \cite{berthelot2019remixmatch}, UDA \cite{xie2020unsupervised}, FixMatch \cite{sohn2020fixmatch}, Dash \cite{xu2021dash}, MPL \cite{pham2021meta}, FlexMatch \cite{zhang2021flexmatch}, SoftMatch \cite{chen2023softmatch}, FreeMatch \cite{wang2023freematch}, SequenceMatch \cite{nguyen2024sequencematch}, and FlatMatch \cite{huang2023flatmatch}. To underscore the rationale behind leveraging Mixup and the effectiveness of RegMixMatch, we compare its training time and classification performance with prior Mixup-based SSL methods \cite{berthelot2019remixmatch} and the latest state-of-the-art SSL approaches \cite{wang2023freematch, huang2023flatmatch}. Additionally, we perform comprehensive ablation studies and hyperparameter analysis to validate the design choices behind RegMixMatch. We also present the results of RegMixMatch on pre-trained backbones in the appendix.

For a fair comparison, and in line with previous SSL studies, we use Wide ResNet-28-2 \cite{zagoruyko2016wide} for CIFAR-10 and SVHN, Wide ResNet-28-8 for CIFAR-100, and ResNet-37-2 \cite{he2016deep} for STL-10. All hyperparameter settings are kept consistent with those used in prior work, as detailed in the appendix. For the implementation of RegMixMatch, the threshold \( \tau_c \) for the consistency loss is set in accordance with the FreeMatch approach, as adopted in the state-of-the-art FlatMatch method. Additionally, we provide experimental results based on the settings of FixMatch in the appendix. The values of \( \tau_m \), \( \alpha_h \), and \( \alpha_l \) are set to 0.999, 1.0, and 16.0, respectively.

\subsection{Main Results}

Table \ref{tab:comparsion} compares the performance of RegMixMatch with existing methods across various datasets. In 12 commonly used SSL scenarios, RegMixMatch achieves state-of-the-art results in 9 cases, demonstrating significant improvement. In the remaining 3 cases, it secures the second-best performance. Notably, previous methods either exhibit minimal improvement or perform poorly in specific scenarios, such as certain datasets or particular labeled data quantities. Our experimental results highlight the comprehensiveness of RegMixMatch. Specifically, RegMixMatch achieves an error rate of only 4.35\% on CIFAR-10 with 10 labels and 11.74\% on STL-10 with 40 labels, surpassing the second-best results by 3.72\% and 3.71\%, respectively.

\begin{table}[h]
	\centering
        \setlength{\tabcolsep}{2pt}
		\begin{tabular}{c|cccc}
			\toprule
			  \# Label &  \multicolumn{1}{c}{FixMatch} & \multicolumn{1}{c}{FlexMatch} &  \multicolumn{1}{c}{FreeMatch} &  \multicolumn{1}{c}{RegMixMatch}  \\
			\cmidrule(r){1-5}
			1w & 46.39 & 45.79 & 45.31 & \textbf{41.65} \\
			10w & 28.47 & 27.83 & 27.43 & \textbf{26.34} \\
			\bottomrule
		\end{tabular}
        \caption{Error rates on ImageNet.} 
        \label{tab:Imagen}
\end{table}


Table \ref{tab:Imagen} presents the results of training RegMixMatch on the ImageNet dataset using MAE pre-trained ViT-B \cite{he2022masked}. RegMixMatch notably outperforms other methods, particularly achieving a 3.66\% improvement on ImageNet with 10,000 labels. More comprehensive results and hyperparameter settings are detailed in the appendix.

\subsection{Efficiency Study}
\label{sec:effi}

\begin{figure}
  \centering
  \includegraphics[width=\linewidth]{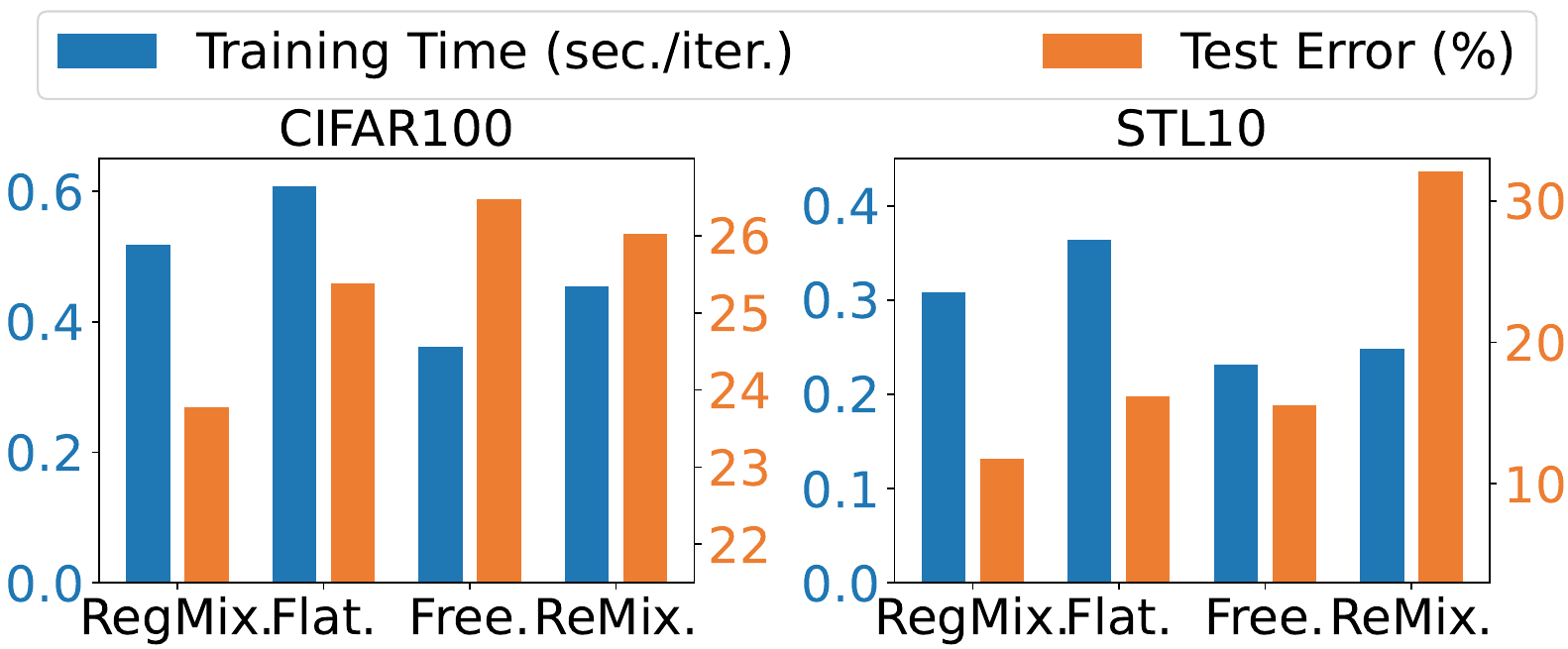}
  \caption{ Efficiency analysis of RegMixMatch.}\label{fig:eff}
\end{figure}

For algorithms handling classification tasks, both runtime and classification performance are critical factors. In this section, we compare RegMixMatch with several SSL algorithms in terms of both metrics. Specifically, we select FlatMatch (the previous state-of-the-art method), FreeMatch (benchmark), and ReMixMatch (Mixup-based SSL method) for comparison. The experiments are conducted on 2 24GB RTX 3090 GPU. Figure \ref{fig:eff} presents a comparison of the 4 algorithms under two settings: CIFAR-100 with 2500 labels and STL-10 with 40 labels. The results show that RegMixMatch not only significantly improves classification performance compared to FlatMatch but also requires less training time, highlighting the efficiency of RegMixMatch. Moreover, compared to ReMixMatch and FreeMatch, RegMixMatch achieves substantial performance gains with only a minimal increase in runtime, demonstrating the effectiveness of leveraging Mixup.

\subsection{Ablation Study}
\label{sec:ablation}


In addition to utilizing low-confidence samples, a key distinction of RegMixMatch from other Mixup-based SSL methods is its retention of clean samples within SRM during training. To verify the rationale behind this design, we compare RegMixMatch with two scenarios: one where only clean samples are retained during training (w/o mixed samples) and another where only mixed samples are used (w/o clean samples). To further validate the rationale and effectiveness of CAM in leveraging low-confidence data, we also present results for two additional configurations: removing CAM entirely (w/o CAM), and replacing CAM with Mixup (CAM $\rightarrow$ Mixup). The results for these ablation studies on STL-10 with 40 labels are shown in Table \ref{tab:ablation}.

\begin{figure}
  \centering
  \captionsetup[subfigure]{font=scriptsize,labelfont=scriptsize}
  \begin{subfigure}{0.49\linewidth}
    \includegraphics[width=\linewidth]{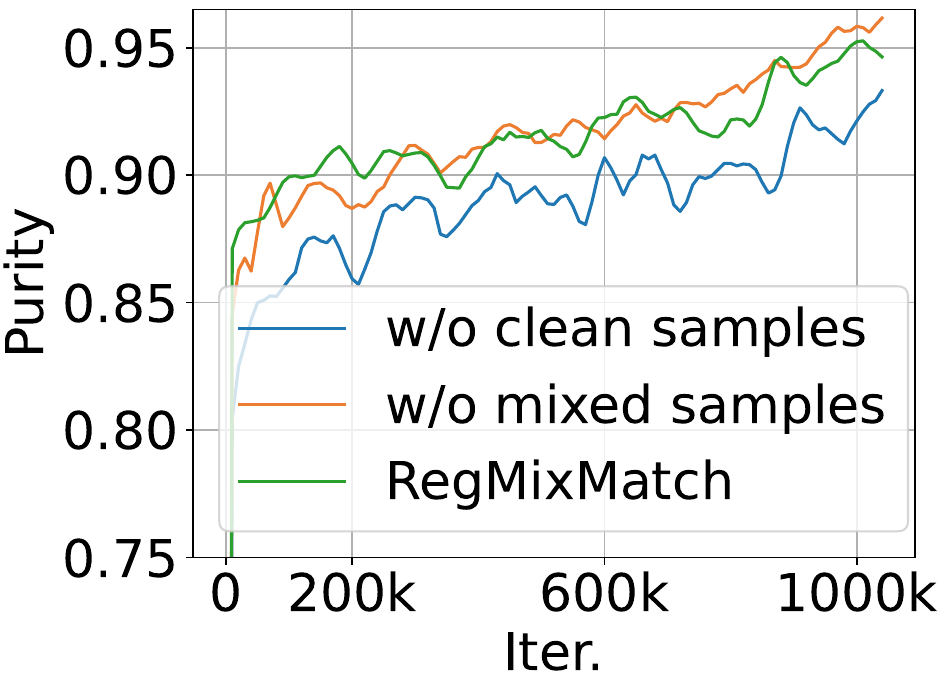}
    \caption{Purity of artificial labels.} \label{fig:free_confi}
  \end{subfigure}%
  \hfill
  \begin{subfigure}{0.49\linewidth}
    \includegraphics[width=\linewidth]{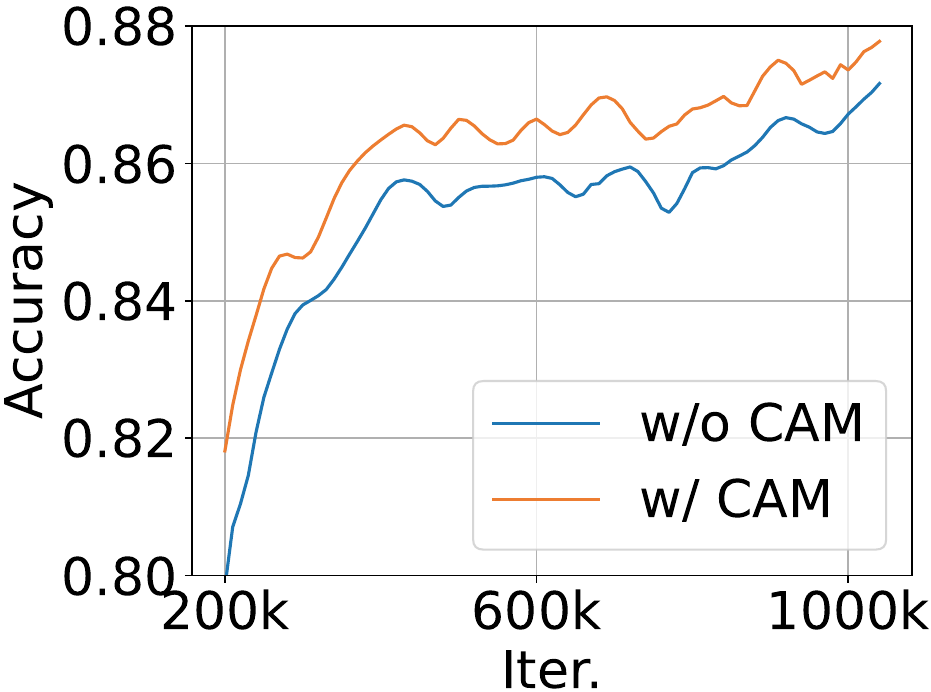}
    \caption{Learning efficiency.} \label{fig:conv_speed}
  \end{subfigure}
  \caption{ Ablation study of RegMixMatch. (a) demonstrates that RegMixMatch alleviates the reduced purity caused by Mixup. (b) depicts CAM improves learning efficiency.}\label{fig:abla}
\end{figure}

Table \ref{tab:ablation} highlights several key findings: 1) Training with only clean samples or only mixed samples yields suboptimal results compared to using both. RegMixMatch leverages SRM not only to enhance generalization performance but also to mitigate the negative effects of high-entropy behavior in SSL, as shown in Figure \ref{fig:abla}(a). 2) CAM improves both classification accuracy and learning efficiency, as evidenced by the higher accuracy at the same iterations in Figure \ref{fig:abla}(b). This suggests that CAM facilitates the safer utilization of low-confidence samples, enriching the training process with more informative data. 3) Replacing CAM with Mixup results in performance degradation.

\begin{table}[b]
    \centering
    \begin{tabular}{c c}
        \toprule
        Ablation  & Avg. Error Rate (\%) \\
        \midrule
        \textbf{RegMixMatch} &  \textbf{11.74} \\
        w/o mixed samples  & 15.56 (+3.82) \\
        w/o clean samples  & 15.78 (+4.04) \\
        w/o CAM  & 12.30 (+0.56) \\
        CAM $\rightarrow$ Mixup & 12.12 (+0.38) \\
        \bottomrule
    \end{tabular}
    \caption{Ablation results of RegMixMatch.} 
    \label{tab:ablation}
\end{table}

\subsection{Sensitivity Analysis}
\label{sec:sensit}

RegMixMatch incorporates three key hyperparameters: the threshold \( \tau_m \), which retains high-confidence samples for SRM; the parameter \( \alpha_h \), which controls the mixing intensity for high-confidence samples; and the parameter \( \alpha_l \), which controls the mixing intensity for low-confidence samples. To assess the specific impact of these hyperparameters on model performance, we conduct experiments on CIFAR-10 with 250 labels.

\begin{figure}
  \centering
  \captionsetup[subfigure]{font=scriptsize,labelfont=scriptsize}
  \begin{subfigure}{0.33\linewidth}
    \includegraphics[width=\linewidth]{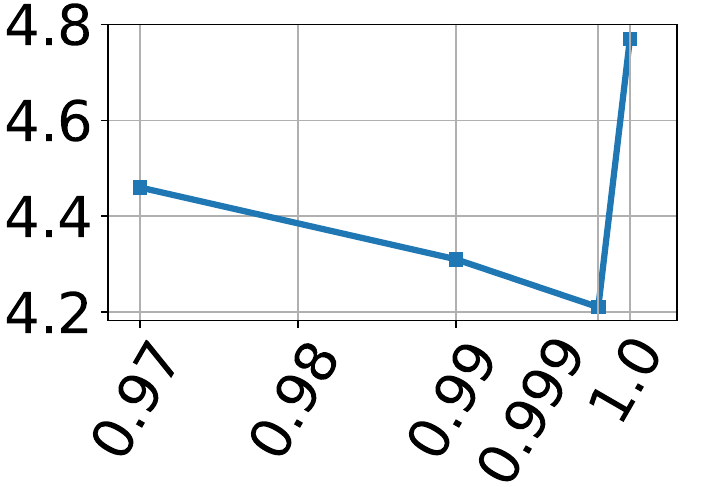}
    \caption{Threshold \( \tau_m \).} \label{fig:tau_m}
  \end{subfigure}%
  \begin{subfigure}{0.32\linewidth}
    \includegraphics[width=\linewidth]{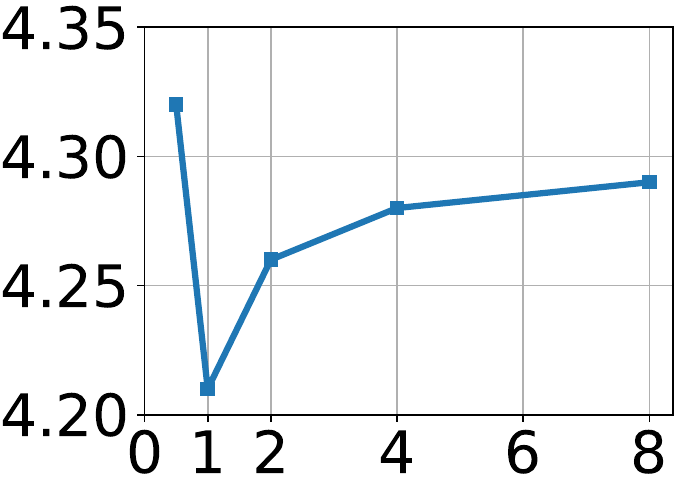}
    \caption{Mixing intensity \( \alpha_h \).} \label{fig:alpha_h}
  \end{subfigure}%
  \hfill
  \begin{subfigure}{0.33\linewidth}
    \includegraphics[width=\linewidth]{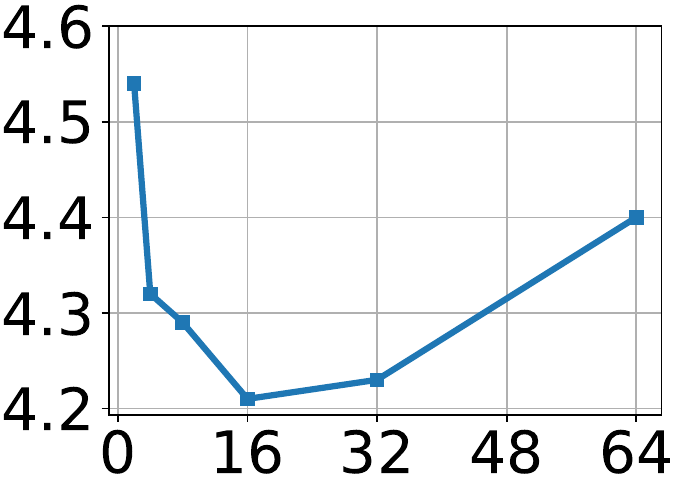}
    \caption{Mixing intensity \( \alpha_l \).} \label{fig:alpha_l}
  \end{subfigure}
  
  \caption{ Parameter sensitivity analysis of RegMixMatch.}\label{fig:sens}
\end{figure}

\paragraph{Confidence Threshold.}
As illustrated in Figure \ref{fig:sens}(a), the optimal value for \( \tau_m \) is 0.999, which is higher than the threshold \( \tau_c \) (0.95). During model training, high-confidence clean samples are used to compute the consistency loss based on pseudo labels, validating the model's predictions on these samples. If \( \tau_m \) is set to a lower value (e.g., 0.95), it would further validate the model's predictions on these samples, leading to overconfidence and exacerbating confirmation bias. Thus, \( \tau_m = 0.999 \) is chosen to balance the quality and quantity of artificial labels for mixed samples.

\paragraph{Mixing Intensity.} 
In addition to retaining clean samples, RegMixMatch distinguishes itself from previous Mixup-based SSL methods by generating more robust mixed samples to enhance generalization. The parameter \( \alpha \) typically set to a value less than 1.0 in prior work. However, as shown in Figures \ref{fig:sens}(b) and \ref{fig:sens}(c), the optimal values for \( \alpha_h \) and \( \alpha_l \) in RegMixMatch are 1.0 and 16.0, respectively. A higher \( \alpha \) indicates stronger augmentation, improving generalization performance. It is noteworthy that \( \alpha_l \gg \alpha_h \). As observed with Mixup, a greater mixing intensity enhances the model’s robustness to noise. Since the supervisory information for low-confidence samples is more prone to noise, a higher \( \alpha_l \) generates more intermediate-state samples, increasing the quantity and variability of the training data, and thereby reducing the risk of overfitting to specific noisy samples. 

\section{Conclusion}
\label{section5}

In this paper, we first demonstrate that Mixup's high-entropy behavior degrades the purity of artificial labels in SSL. To address this issue, we introduce SRM, a framework that integrates pseudo-labeling with consistency regularization through the application of RegMixup in SSL. We show that SRM effectively mitigates the reduced artificial labels purity, enhancing Mixup's utility for high-confidence samples. Additionally, we investigate strategies for leveraging low-confidence samples. Specifically, we propose CAM, a method that combines low-confidence samples with specific class samples to enhance the quality and utility of artificial labels. Extensive experiments validate the efficiency and effectiveness of RegMixMatch. Our approach represents an initial exploration into the potential of low-confidence samples, with further in-depth studies recommended as a direction for future research.

\newpage

\section{Acknowledgments}
This work was supported by the National Natural Science Foundation of China (No. 62402031) and the Beijing Nova Program (No. 20240484620).

\bibliography{aaai25}

\begin{thebibliography}{49}
\providecommand{\natexlab}[1]{#1}

\bibitem[{Bachman, Alsharif, and Precup(2014)}]{bachman2014learning}
Bachman, P.; Alsharif, O.; and Precup, D. 2014.
\newblock Learning with pseudo-ensembles.
\newblock \emph{Advances in neural information processing systems}, 27.

\bibitem[{Bengio et~al.(2009)Bengio, Louradour, Collobert, and Weston}]{bengio2009curriculum}
Bengio, Y.; Louradour, J.; Collobert, R.; and Weston, J. 2009.
\newblock Curriculum learning.
\newblock In \emph{Proceedings of the 26th annual international conference on machine learning}, 41--48.

\bibitem[{Berthelot et~al.(2019{\natexlab{a}})Berthelot, Carlini, Cubuk, Kurakin, Sohn, Zhang, and Raffel}]{berthelot2019remixmatch}
Berthelot, D.; Carlini, N.; Cubuk, E.~D.; Kurakin, A.; Sohn, K.; Zhang, H.; and Raffel, C. 2019{\natexlab{a}}.
\newblock Remixmatch: Semi-supervised learning with distribution alignment and augmentation anchoring.
\newblock \emph{arXiv preprint arXiv:1911.09785}.

\bibitem[{Berthelot et~al.(2019{\natexlab{b}})Berthelot, Carlini, Goodfellow, Papernot, Oliver, and Raffel}]{berthelot2019mixmatch}
Berthelot, D.; Carlini, N.; Goodfellow, I.; Papernot, N.; Oliver, A.; and Raffel, C.~A. 2019{\natexlab{b}}.
\newblock Mixmatch: A holistic approach to semi-supervised learning.
\newblock \emph{Advances in neural information processing systems}, 32.

\bibitem[{Chapelle et~al.(2000)Chapelle, Weston, Bottou, and Vapnik}]{chapelle2000vicinal}
Chapelle, O.; Weston, J.; Bottou, L.; and Vapnik, V. 2000.
\newblock Vicinal risk minimization.
\newblock \emph{Advances in neural information processing systems}, 13.

\bibitem[{Chen et~al.(2023)Chen, Tao, Fan, Wang, Wang, Schiele, Xie, Raj, and Savvides}]{chen2023softmatch}
Chen, H.; Tao, R.; Fan, Y.; Wang, Y.; Wang, J.; Schiele, B.; Xie, X.; Raj, B.; and Savvides, M. 2023.
\newblock SoftMatch: Addressing the quantity-quality tradeoff in semi-supervised learning.
\newblock In \emph{The eleventh international conference on learning representations}.

\bibitem[{Chen et~al.(2020)Chen, Kornblith, Norouzi, and Hinton}]{chen2020simple}
Chen, T.; Kornblith, S.; Norouzi, M.; and Hinton, G. 2020.
\newblock A simple framework for contrastive learning of visual representations.
\newblock In \emph{International conference on machine learning}, 1597--1607. PMLR.

\bibitem[{Coates, Ng, and Lee(2011)}]{coates2011analysis}
Coates, A.; Ng, A.; and Lee, H. 2011.
\newblock An analysis of single-layer networks in unsupervised feature learning.
\newblock In \emph{Proceedings of the fourteenth international conference on artificial intelligence and statistics}, 215--223.

\bibitem[{Cubuk et~al.(2018)Cubuk, Zoph, Mane, Vasudevan, and Le}]{cubuk2018autoaugment}
Cubuk, E.~D.; Zoph, B.; Mane, D.; Vasudevan, V.; and Le, Q.~V. 2018.
\newblock Autoaugment: Learning augmentation policies from data.
\newblock \emph{arXiv preprint arXiv:1805.09501}.

\bibitem[{Cubuk et~al.(2019)Cubuk, Zoph, Mane, Vasudevan, and Le}]{cubuk2019autoaugment}
Cubuk, E.~D.; Zoph, B.; Mane, D.; Vasudevan, V.; and Le, Q.~V. 2019.
\newblock Autoaugment: Learning augmentation strategies from data.
\newblock In \emph{Proceedings of the IEEE/CVF conference on computer vision and pattern recognition}, 113--123.

\bibitem[{Cubuk et~al.(2020)Cubuk, Zoph, Shlens, and Le}]{cubuk2020randaugment}
Cubuk, E.~D.; Zoph, B.; Shlens, J.; and Le, Q.~V. 2020.
\newblock Randaugment: Practical automated data augmentation with a reduced search space.
\newblock In \emph{Proceedings of the IEEE/CVF conference on computer vision and pattern recognition workshops}, 702--703.

\bibitem[{Deng et~al.(2009)Deng, Dong, Socher, Li, Li, and Fei-Fei}]{deng2009imagenet}
Deng, J.; Dong, W.; Socher, R.; Li, L.-J.; Li, K.; and Fei-Fei, L. 2009.
\newblock Imagenet: A large-scale hierarchical image database.
\newblock In \emph{2009 IEEE conference on computer vision and pattern recognition}, 248--255. Ieee.

\bibitem[{DeVries and Taylor(2017)}]{devries2017improved}
DeVries, T.; and Taylor, G.~W. 2017.
\newblock Improved regularization of convolutional neural networks with cutout.
\newblock \emph{arXiv preprint arXiv:1708.04552}.

\bibitem[{Dosovitskiy(2020)}]{dosovitskiy2020image}
Dosovitskiy, A. 2020.
\newblock An image is worth 16x16 words: Transformers for image recognition at scale.
\newblock \emph{arXiv preprint arXiv:2010.11929}.

\bibitem[{Goodfellow, Shlens, and Szegedy(2014)}]{goodfellow2014explaining}
Goodfellow, I.~J.; Shlens, J.; and Szegedy, C. 2014.
\newblock Explaining and harnessing adversarial examples.
\newblock \emph{arXiv preprint arXiv:1412.6572}.

\bibitem[{He et~al.(2022)He, Chen, Xie, Li, Doll{\'a}r, and Girshick}]{he2022masked}
He, K.; Chen, X.; Xie, S.; Li, Y.; Doll{\'a}r, P.; and Girshick, R. 2022.
\newblock Masked autoencoders are scalable vision learners.
\newblock In \emph{Proceedings of the IEEE/CVF conference on computer vision and pattern recognition}, 16000--16009.

\bibitem[{He et~al.(2016)He, Zhang, Ren, and Sun}]{he2016deep}
He, K.; Zhang, X.; Ren, S.; and Sun, J. 2016.
\newblock Deep residual learning for image recognition.
\newblock In \emph{Proceedings of the IEEE conference on computer vision and pattern recognition}, 770--778.

\bibitem[{Helber et~al.(2019)Helber, Bischke, Dengel, and Borth}]{helber2019eurosat}
Helber, P.; Bischke, B.; Dengel, A.; and Borth, D. 2019.
\newblock Eurosat: A novel dataset and deep learning benchmark for land use and land cover classification.
\newblock \emph{IEEE journal of selected topics in applied earth observations and remote sensing}, 12(7): 2217--2226.

\bibitem[{Hong, Choi, and Kim(2021)}]{hong2021stylemix}
Hong, M.; Choi, J.; and Kim, G. 2021.
\newblock Stylemix: Separating content and style for enhanced data augmentation.
\newblock In \emph{Proceedings of the IEEE/CVF conference on computer vision and pattern recognition}, 14862--14870.

\bibitem[{Huang et~al.(2023)Huang, Shen, Yu, Han, and Liu}]{huang2023flatmatch}
Huang, Z.; Shen, L.; Yu, J.; Han, B.; and Liu, T. 2023.
\newblock Flatmatch: Bridging labeled data and unlabeled data with cross-sharpness for semi-supervised learning.
\newblock \emph{Advances in neural information processing systems}, 36: 18474--18494.

\bibitem[{Krizhevsky, Hinton et~al.(2009)}]{krizhevsky2009learning}
Krizhevsky, A.; Hinton, G.; et~al. 2009.
\newblock Learning multiple layers of features from tiny images.

\bibitem[{Laine and Aila(2017)}]{laine2017temporal}
Laine, S.; and Aila, T. 2017.
\newblock Temporal ensembling for semi-supervised Learning.
\newblock In \emph{International conference on learning representations}.

\bibitem[{Lee et~al.(2013)}]{lee2013pseudo}
Lee, D.-H.; et~al. 2013.
\newblock Pseudo-label: The simple and efficient semi-supervised learning method for deep neural networks.
\newblock In \emph{Workshop on challenges in representation learning, ICML}, volume~3, 896. Atlanta.

\bibitem[{Li, Xiong, and Hoi(2021)}]{li2021comatch}
Li, J.; Xiong, C.; and Hoi, S.~C. 2021.
\newblock Comatch: Semi-supervised learning with contrastive graph regularization.
\newblock In \emph{Proceedings of the IEEE/CVF international conference on computer vision}, 9475--9484.

\bibitem[{Li et~al.(2023)Li, Wang, Liu, Wu, and Li}]{li2023openmixup}
Li, S.; Wang, Z.; Liu, Z.; Wu, D.; and Li, S.~Z. 2023.
\newblock Openmixup: Open mixup toolbox and benchmark for visual representation learning.
\newblock \emph{arXiv preprint arXiv:2209.04851}.

\bibitem[{Liu et~al.(2022)Liu, Zeng, Chen, Xu, Lai, Ma, and Xu}]{liu2022scinet}
Liu, M.; Zeng, A.; Chen, M.; Xu, Z.; Lai, Q.; Ma, L.; and Xu, Q. 2022.
\newblock Scinet: Time series modeling and forecasting with sample convolution and interaction.
\newblock \emph{Advances in Neural Information Processing Systems}, 35: 5816--5828.

\bibitem[{Miyato et~al.(2018)Miyato, Maeda, Koyama, and Ishii}]{miyato2018virtual}
Miyato, T.; Maeda, S.-i.; Koyama, M.; and Ishii, S. 2018.
\newblock Virtual adversarial training: a regularization method for supervised and semi-supervised learning.
\newblock \emph{IEEE transactions on pattern analysis and machine intelligence}, 41(8): 1979--1993.

\bibitem[{Netzer et~al.(2011)Netzer, Wang, Coates, Bissacco, Wu, Ng et~al.}]{netzer2011reading}
Netzer, Y.; Wang, T.; Coates, A.; Bissacco, A.; Wu, B.; Ng, A.~Y.; et~al. 2011.
\newblock Reading digits in natural images with unsupervised feature learning.
\newblock In \emph{NIPS workshop on deep learning and unsupervised feature learning}, volume 2011, 4. Granada.

\bibitem[{Nguyen(2024)}]{nguyen2024sequencematch}
Nguyen, K.-B. 2024.
\newblock SequenceMatch: Revisiting the design of weak-strong augmentations for semi-supervised learning.
\newblock In \emph{Proceedings of the IEEE/CVF winter conference on applications of computer vision}, 96--106.

\bibitem[{Pham et~al.(2021)Pham, Dai, Xie, and Le}]{pham2021meta}
Pham, H.; Dai, Z.; Xie, Q.; and Le, Q.~V. 2021.
\newblock Meta pseudo labels.
\newblock In \emph{Proceedings of the IEEE/CVF conference on computer vision and pattern recognition}, 11557--11568.

\bibitem[{Pinto et~al.(2022)Pinto, Yang, Lim, Torr, and Dokania}]{pinto2022using}
Pinto, F.; Yang, H.; Lim, S.~N.; Torr, P.; and Dokania, P. 2022.
\newblock Using mixup as a regularizer can surprisingly improve accuracy \& out-of-distribution robustness.
\newblock \emph{Advances in neural information processing systems}, 35: 14608--14622.

\bibitem[{Qin et~al.(2020)Qin, Fang, Zhang, Liu, Wang, and Wang}]{qin2020resizemix}
Qin, J.; Fang, J.; Zhang, Q.; Liu, W.; Wang, X.; and Wang, X. 2020.
\newblock Resizemix: Mixing data with preserved object information and true labels.
\newblock \emph{arXiv preprint arXiv:2012.11101}.

\bibitem[{Rasmus et~al.(2015)Rasmus, Berglund, Honkala, Valpola, and Raiko}]{rasmus2015semi}
Rasmus, A.; Berglund, M.; Honkala, M.; Valpola, H.; and Raiko, T. 2015.
\newblock Semi-supervised learning with ladder networks.
\newblock \emph{Advances in neural information processing systems}, 28.

\bibitem[{Sohn et~al.(2020)Sohn, Berthelot, Carlini, Zhang, Zhang, Raffel, Cubuk, Kurakin, and Li}]{sohn2020fixmatch}
Sohn, K.; Berthelot, D.; Carlini, N.; Zhang, Z.; Zhang, H.; Raffel, C.~A.; Cubuk, E.~D.; Kurakin, A.; and Li, C.-L. 2020.
\newblock Fixmatch: Simplifying semi-supervised learning with consistency and confidence.
\newblock \emph{Advances in neural information processing systems}, 33: 596--608.

\bibitem[{Tarvainen and Valpola(2017)}]{tarvainen2017mean}
Tarvainen, A.; and Valpola, H. 2017.
\newblock Mean teachers are better role models: Weight-averaged consistency targets improve semi-supervised deep learning results.
\newblock \emph{Advances in neural information processing systems}, 30.

\bibitem[{Uddin et~al.(2021)Uddin, Monira, Shin, Chung, and Bae}]{uddin2021saliencymix}
Uddin, A. F. M.~S.; Monira, M.~S.; Shin, W.; Chung, T.; and Bae, S.-H. 2021.
\newblock SaliencyMix: A saliency guided data augmentation strategy for better regularization.
\newblock In \emph{International conference on learning representations}.

\bibitem[{Vapnik(1991)}]{vapnik1991principles}
Vapnik, V. 1991.
\newblock Principles of risk minimization for learning theory.
\newblock \emph{Advances in neural information processing systems}, 4.

\bibitem[{Verma et~al.(2019)Verma, Lamb, Beckham, Najafi, Mitliagkas, Lopez-Paz, and Bengio}]{verma2019manifold}
Verma, V.; Lamb, A.; Beckham, C.; Najafi, A.; Mitliagkas, I.; Lopez-Paz, D.; and Bengio, Y. 2019.
\newblock Manifold mixup: Better representations by interpolating hidden states.
\newblock In \emph{International conference on machine learning}, 6438--6447. PMLR.

\bibitem[{Wang et~al.(2022)Wang, Chen, Fan, Sun, Tao, Hou, Wang, Yang, Zhou, Guo et~al.}]{wang2022usb}
Wang, Y.; Chen, H.; Fan, Y.; Sun, W.; Tao, R.; Hou, W.; Wang, R.; Yang, L.; Zhou, Z.; Guo, L.-Z.; et~al. 2022.
\newblock Usb: A unified semi-supervised learning benchmark for classification.
\newblock \emph{Advances in neural information processing systems}, 35: 3938--3961.

\bibitem[{Wang et~al.(2023)Wang, Chen, Heng, Hou, Fan, Wu, Wang, Savvides, Shinozaki, Raj, Schiele, and Xie}]{wang2023freematch}
Wang, Y.; Chen, H.; Heng, Q.; Hou, W.; Fan, Y.; Wu, Z.; Wang, J.; Savvides, M.; Shinozaki, T.; Raj, B.; Schiele, B.; and Xie, X. 2023.
\newblock FreeMatch: Self-adaptive thresholding for semi-supervised learning.
\newblock In \emph{The eleventh international conference on learning representations}.

\bibitem[{Wei et~al.(2023)Wei, Wang, Yuan, Li, and Chen}]{wei2023time}
Wei, C.; Wang, Z.; Yuan, J.; Li, C.; and Chen, S. 2023.
\newblock Time-frequency based multi-task learning for semi-supervised time series classification.
\newblock \emph{Information sciences}, 619: 762--780.

\bibitem[{Wen et~al.(2021)Wen, Jerfel, Muller, Dusenberry, Snoek, Lakshminarayanan, and Tran}]{wen2021combining}
Wen, Y.; Jerfel, G.; Muller, R.; Dusenberry, M.~W.; Snoek, J.; Lakshminarayanan, B.; and Tran, D. 2021.
\newblock Combining ensembles and data augmentation can harm your calibration.
\newblock In \emph{International conference on learning representations}.

\bibitem[{Xie et~al.(2020)Xie, Dai, Hovy, Luong, and Le}]{xie2020unsupervised}
Xie, Q.; Dai, Z.; Hovy, E.; Luong, T.; and Le, Q. 2020.
\newblock Unsupervised data augmentation for consistency training.
\newblock \emph{Advances in neural information processing systems}, 33: 6256--6268.

\bibitem[{Xu et~al.(2021)Xu, Shang, Ye, Qian, Li, Sun, Li, and Jin}]{xu2021dash}
Xu, Y.; Shang, L.; Ye, J.; Qian, Q.; Li, Y.-F.; Sun, B.; Li, H.; and Jin, R. 2021.
\newblock Dash: Semi-supervised learning with dynamic thresholding.
\newblock In \emph{International conference on machine learning}, 11525--11536. PMLR.

\bibitem[{Yun et~al.(2019)Yun, Han, Oh, Chun, Choe, and Yoo}]{yun2019cutmix}
Yun, S.; Han, D.; Oh, S.~J.; Chun, S.; Choe, J.; and Yoo, Y. 2019.
\newblock Cutmix: Regularization strategy to train strong classifiers with localizable features.
\newblock In \emph{Proceedings of the IEEE/CVF international conference on computer vision}, 6023--6032.

\bibitem[{Zagoruyko and Komodakis(2016)}]{zagoruyko2016wide}
Zagoruyko, S.; and Komodakis, N. 2016.
\newblock Wide residual networks.
\newblock \emph{arXiv preprint arXiv:1605.07146}.

\bibitem[{Zhang et~al.(2021)Zhang, Wang, Hou, Wu, Wang, Okumura, and Shinozaki}]{zhang2021flexmatch}
Zhang, B.; Wang, Y.; Hou, W.; Wu, H.; Wang, J.; Okumura, M.; and Shinozaki, T. 2021.
\newblock Flexmatch: Boosting semi-supervised learning with curriculum pseudo labeling.
\newblock \emph{Advances in neural information processing systems}, 34: 18408--18419.

\bibitem[{Zhang et~al.(2018)Zhang, Cisse, Dauphin, and Lopez-Paz}]{zhang2018mixup}
Zhang, H.; Cisse, M.; Dauphin, Y.~N.; and Lopez-Paz, D. 2018.
\newblock mixup: Beyond empirical risk minimization.
\newblock In \emph{International conference on learning representations}.

\bibitem[{Zheng et~al.(2022)Zheng, You, Huang, Wang, Qian, and Xu}]{zheng2022simmatch}
Zheng, M.; You, S.; Huang, L.; Wang, F.; Qian, C.; and Xu, C. 2022.
\newblock Simmatch: Semi-supervised learning with similarity matching.
\newblock In \emph{Proceedings of the IEEE/CVF conference on computer vision and pattern recognition}, 14471--14481.

\end{thebibliography}

\clearpage

\twocolumn[{%
  \newsavebox{\appendixarea}
  \sbox{\appendixarea}{
    \vbox{%
      \hsize\textwidth%
      \linewidth\hsize%
      \vskip 0.625in minus 0.125in %
      \centering%
      {\LARGE\bf Appendix for\\ ``RegMixMatch: Optimizing Mixup Utilization in Semi-Supervised Learning'' \par}%
      \vskip 0.1in plus 0.5fil minus 0.05in %
    }%
  }%
  \usebox{\appendixarea}%
  \vskip 0.5em plus 2fil%
}]

\appendix
In this appendix, we provide experimental details of RegMixMatch for reproduction. Additionally, we include supplementary experimental results to verify the superiority of RegMixMatch in terms of artificial label purity and learning efficiency. Finally, we demonstrate the performance of RegMixMatch in two scenarios: large-scale dataset, and pre-trained backbones.
\section{Experimental Details}
In this section, we present detailed experimental settings and our modifications to ResizeMix \cite{qin2020resizemix}. We also demonstrate the key differences between RegMixMatch and other Mixup \cite{zhang2018mixup} based semi-supervised learning methods \cite{berthelot2019mixmatch, berthelot2019remixmatch}.
\subsection{Experimental Setting}
Followed FreeMatch \cite{wang2023freematch}, we categorize the hyperparameter settings into algorithm-dependent and algorithm-independent groups, and present them in Table \ref{tab:alg_dep} and Table \ref{tab:alg_ind}, respectively.

\begin{table}[!htbp]
	\centering
        \setlength{\tabcolsep}{0.5pt}
		\begin{tabular}{cccccc}
			\toprule
			Algorithm &  RegMixMatch \\\cmidrule(r){1-1} \cmidrule(lr){2-2}
			Unlabeled data to labeled data ratio    &  7 \\
			\cmidrule(r){1-1}\cmidrule(lr){2-2}
			Threshold $\tau_m$ for implementing Mixup   &  0.999 \\
			\cmidrule(r){1-1}\cmidrule(lr){2-2}
			Mixing intensity $\alpha_h$ for high-confidence data   &  1.0 \\
			\cmidrule(r){1-1}\cmidrule(lr){2-2}
			Mixing intensity $\alpha_l$ for low-confidence data   &  16.0 \\
			\cmidrule(r){1-1}\cmidrule(lr){2-2}
			Thresholding EMA decay  & 0.999 \\
			\bottomrule
		\end{tabular}
	\caption{Algorithm-dependent hyperparameters.}
	\label{tab:alg_dep}
\end{table}

\begin{table}[!htbp]
	\centering
        \setlength{\tabcolsep}{1pt}

		\begin{tabular}{ccccc}\toprule
			Dataset &  CIFAR-10 & CIFAR-100 & STL-10 & SVHN  \\\cmidrule(r){1-1} \cmidrule(lr){2-2}\cmidrule(lr){3-3}\cmidrule(lr){4-4}\cmidrule(l){5-5}
			Model    &  WRN-28 & WRN-28 & WRN-37 & WRN-28  \\    &  -2 & -8 & -2 & -2\\\cmidrule(r){1-1} \cmidrule(lr){2-2}\cmidrule(lr){3-3}\cmidrule(lr){4-4}\cmidrule(l){5-5}
			Weight decay&  5e-4  & 1e-3 & 5e-4 & 5e-4\\ \cmidrule(r){1-1} \cmidrule(lr){2-2}\cmidrule(lr){3-3}\cmidrule(lr){4-4}\cmidrule(l){5-5}
			Batch size & \multicolumn{4}{c}{64} \\\cmidrule(r){1-1} \cmidrule(l){2-5}
			Learning rate & \multicolumn{4}{c}{0.03}\\\cmidrule(r){1-1} \cmidrule(l){2-5}
			SGD momentum & \multicolumn{4}{c}{0.9}\\\cmidrule(r){1-1} \cmidrule(l){2-5}
			EMA decay & \multicolumn{4}{c}{0.999}\\
			\bottomrule
		\end{tabular}
	\caption{Algorithm-independent hyperparameters.}
	\label{tab:alg_ind}
\end{table}

\subsection{ResizeMix}
For reproduction, we show our modifications to ResizeMix. ResizeMix resizes the source image through a scaling factor \(\tau\) to a smaller patch \(P\), where \(\tau \sim U(0.1, 0.8)\). The resized patch is then pasted onto a random region of the target image. In this way, the mixing ratio \(\lambda\) can be expressed using the area ratio of the patch to the target image as follows:

\begin{equation}
\lambda = \frac{W_p \times H_p}{W \times H},
\end{equation}
where \(W\), \(H\), \(W_p\), and \(H_p\) represent the width and height of the target image and the source patch, respectively. We can observe that \(\lambda\) and \(\tau\) satisfy

\begin{equation}
\lambda = \tau^2.
\end{equation}
To adjust the average mixing intensity, we change the distribution of $\tau$ from uniform to a Beta distribution with parameters $(\alpha, \alpha)$. Then, we use the square root of $\tau$ as the scaling factor to resize the images. Consequently, the mixing ratio $\lambda$ follows a Beta distribution with parameters $(\alpha, \alpha)$. The larger the \( \alpha \), the closer \( \lambda \) is to 0.5, resulting in stronger mixing. Finally, similar to Mixup, we can control the mixing intensity in ResizeMix using the $\alpha$ parameter.

\subsection{Key Improvements}
MixMatch \cite{berthelot2019mixmatch} and ReMixMatch \cite{berthelot2019remixmatch} both use Mixup technique to achieve consistency regularization. These methods generate artificial labels using the softmax (or sharpened softmax) outputs for all unlabeled samples, and they are trained using \( \ell \)-2 loss. To better distinguish these methods from RegMixMatch, we list the key improvements RegMixMatch introduces when employing Mixup:
\begin{itemize}
\item We categorize data based on confidence levels, with distinct processing methods applied to each category.
\item For high-confidence data, training is done using pseudo labels and cross-entropy loss, while for low-confidence data, the softmax outputs and \( \ell \)-2 loss are used.
\item The original, non-Mixup samples are retained for training to alleviate the issue of artificial labels purity degradation caused by Mixup.
\item RegMixMatch allows for stronger mixing intensities: the mixing intensity is set to 1.0 for high-confidence data and 16.0 for low-confidence data. The mixing intensity in the other two methods is typically less than 1.0.
\end{itemize}

\section{Additional Experiment Results}
In this section, we compare the performance of RegMixMatch with other SSL algorithms in terms of purity and learning efficiency. Additionally, we present the results of applying other thresholding and mixing strategy to RegMixMatch.

\subsection{Purity Comparison}
\begin{figure}[!htbp]
  \centering
  \includegraphics[width=\linewidth]{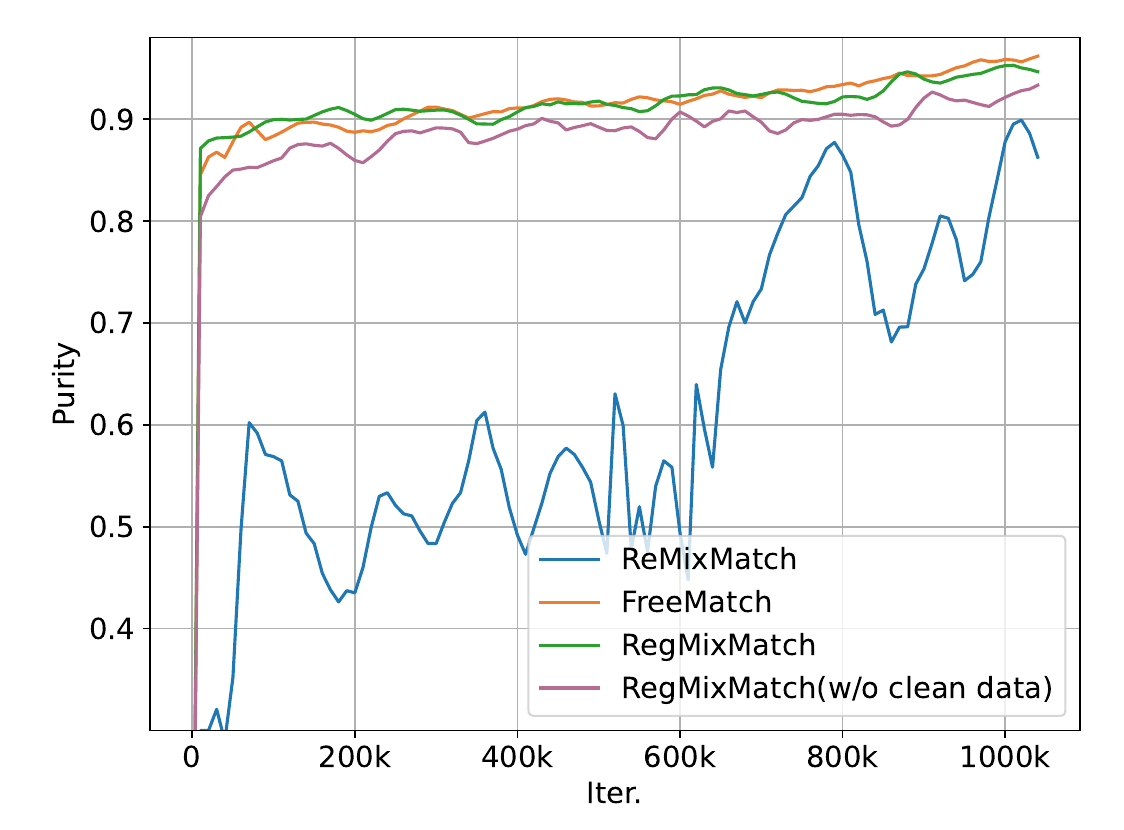}
  \caption{ Purity of different algorithms.}\label{fig:purity}
\end{figure}

In Figure \ref{fig:purity} we presented the changes in purity during the training process of different SSL algorithms. We can observe that although ReMixMatch enhances the purity of artificial labels by sharpening the softmax outputs, the introduction of Mixup technique results in consistently lower and more fluctuating purity of the artificial labels during training. FreeMatch \cite{wang2023freematch} significantly improves the purity of artificial labels by utilizing clean samples for training through the pseudo-labeling technique. Our method, RegMixMatch, while incorporating the Mixup technique, retains the training on clean samples, effectively mitigating the issue of reduced label purity caused by Mixup and achieving nearly the same level of purity as FreeMatch.

\begin{figure}[!htbp]
  \centering
  \includegraphics[width=\linewidth]{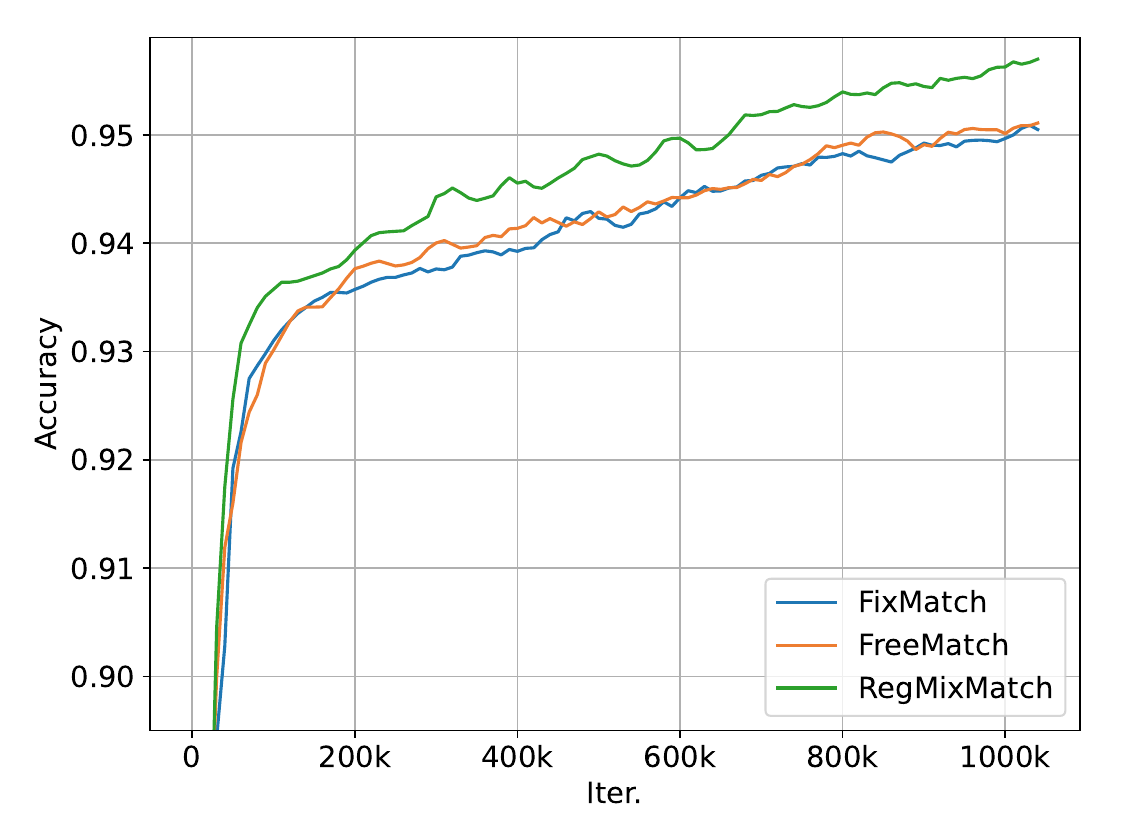}
  \caption{ Learning efficiency of different algorithms.}\label{fig:algo_learneff}
\end{figure}

\subsection{Learning Efficiency Comparison}
We would like to emphasize the superior learning efficiency of RegMixMatch. We presented the changes in accuracy of the RegMixMatch, FixMatch \cite{sohn2020fixmatch}, and FreeMatch algorithms during the training process on CIFAR-10 with 250 labels. From Figure \ref{fig:algo_learneff}, we can observe that by effectively leveraging both high and low-confidence samples, RegMixMatch achieves higher accuracy compared to other SSL algorithms within the same number of iterations.

\subsection{Quantitative Results}
As introduced in the paper, RegMixMatch adopts dynamic threshold strategy proposed by FreeMatch for the setting of \( \tau_c \) and uses ResizeMix technique to generate mixed images. We conduct additional experiments on 1) using fixed threshold strategy \cite{sohn2020fixmatch} for \( \tau_c \). 2) using Mixup technique for image mixing. The results for these studies on CIFAR-10 are shown in Table \ref{tab:thresholdst} and Table \ref{tab:mixst}. We can see that our method demonstrates superior classification performance under both threshold strategies, and the combination of the dynamic threshold strategy with the ResizeMix image augmentation technique achieves the optimal results.

\begin{table}[t]
	\centering
		\begin{tabular}{l|ccc}
			\toprule
			Dataset &  \multicolumn{3}{c}{CIFAR10} \\
			\cmidrule(r){1-1} \cmidrule(l){2-4}
			\# Label   &  \multicolumn{1}{c}{40} & \multicolumn{1}{c}{250} & \multicolumn{1}{c}{4000}\\ \cmidrule{1-4}
			FixMatch & 7.47{\scriptsize $\pm$0.28} & 4.86{\scriptsize $\pm$0.05} & 4.21{\scriptsize $\pm$0.08} \\
			RegMixMatch (Fix) & \textbf{6.02}{\scriptsize $\pm$0.10} & \textbf{4.21}{\scriptsize $\pm$0.03} & \textbf{3.73}{\scriptsize $\pm$0.05} \\		
			\midrule
			FreeMatch & 4.90{\scriptsize $\pm$0.04} & 4.88{\scriptsize $\pm$0.18} & 4.10{\scriptsize $\pm$0.02} \\
			RegMixMatch (Free) & \textbf{4.24}{\scriptsize $\pm$0.02} & \textbf{4.21}{\scriptsize $\pm$0.02} & \textbf{3.38}{\scriptsize $\pm$0.05} \\
			\bottomrule
		\end{tabular}
	\caption{Performance on other thresholding strategy.}
	\label{tab:thresholdst}
\end{table}

\begin{table}[t]
	\centering
        \setlength{\tabcolsep}{4pt} 
		\begin{tabular}{l|ccc}
			\toprule
			Dataset &  \multicolumn{3}{c}{CIFAR10} \\
			\cmidrule(r){1-1} \cmidrule(l){2-4}
			\# Label   &  \multicolumn{1}{c}{40} & \multicolumn{1}{c}{250} & \multicolumn{1}{c}{4000}\\ \cmidrule{1-4}
			RegMixMatch (Mixup) & 4.40{\scriptsize $\pm$0.03} & 4.35{\scriptsize $\pm$0.02} & 3.62{\scriptsize $\pm$0.06} \\	    
			RegMixMatch & \textbf{4.24}{\scriptsize $\pm$0.02} & \textbf{4.21}{\scriptsize $\pm$0.02} & \textbf{3.38}{\scriptsize $\pm$0.05} \\
			\bottomrule
		\end{tabular}
	\caption{Performance on other mixing strategy.}
	\label{tab:mixst}
\end{table}

\subsection{Pre-trained Backbones}

\begin{table*}[t]
	\centering
	\resizebox{\textwidth}{!}{%
		\begin{tabular}{l|cc|cc|cc}
			\toprule
			Dataset & \multicolumn{2}{c|}{CIFAR100}& \multicolumn{2}{c|}{STL10}& \multicolumn{2}{c}{EuroSAT} \\ \cmidrule(r){1-1}\cmidrule(lr){2-3}\cmidrule(lr){4-5}\cmidrule(l){6-7}
			
			\# Label & \multicolumn{1}{c}{200} & \multicolumn{1}{c|}{400} & \multicolumn{1}{c}{40}  & \multicolumn{1}{c|}{100} & \multicolumn{1}{c}{20}  & \multicolumn{1}{c}{40}\\ \cmidrule(r){1-1}\cmidrule(lr){2-3}\cmidrule(lr){4-5}\cmidrule(l){6-7}
			
			VAT \cite{miyato2018virtual} & 31.49{\scriptsize $\pm$1.33} & 21.34{\scriptsize $\pm$0.50} & 18.45{\scriptsize $\pm$1.47} & 10.69{\scriptsize $\pm$0.51} & 26.16{\scriptsize $\pm$0.96} & 10.09{\scriptsize $\pm$0.94} \\
			Mean Teacher \cite{tarvainen2017mean} & 35.47{\scriptsize $\pm$0.40} & 26.03{\scriptsize $\pm$0.30} & 18.67{\scriptsize $\pm$1.69} & 24.19{\scriptsize $\pm$10.15} & 26.83{\scriptsize $\pm$1.46} & 15.85{\scriptsize $\pm$1.66} \\
			MixMatch \cite{berthelot2019mixmatch} & 38.22{\scriptsize $\pm$0.71} & 26.72{\scriptsize $\pm$0.72} & 58.77{\scriptsize $\pm$1.98} & 36.74{\scriptsize $\pm$1.24} & 24.85{\scriptsize $\pm$4.85} & 17.28{\scriptsize $\pm$2.67} \\
			ReMixMatch \cite{berthelot2019remixmatch} & 22.21{\scriptsize $\pm$2.21} & 16.86{\scriptsize $\pm$0.57} & 13.08{\scriptsize $\pm$3.34} & \underline{7.21}{\scriptsize $\pm$0.39} & \underline{5.05}{\scriptsize $\pm$1.05} & 5.07{\scriptsize $\pm$0.56} \\
			UDA \cite{xie2020unsupervised} & 28.80{\scriptsize $\pm$0.61} & 19.00{\scriptsize $\pm$0.79} & 15.58{\scriptsize $\pm$3.16} & 7.65{\scriptsize $\pm$1.11} & 9.83{\scriptsize $\pm$2.15} & 6.22{\scriptsize $\pm$1.36} \\
			FixMatch \cite{sohn2020fixmatch} & 29.60{\scriptsize $\pm$0.90} & 19.56{\scriptsize $\pm$0.52} & 16.15{\scriptsize $\pm$1.89} & 8.11{\scriptsize $\pm$0.68} & 13.44{\scriptsize $\pm$3.53} & 5.91{\scriptsize $\pm$2.02} \\
			FlexMatch \cite{zhang2021flexmatch} & 26.76{\scriptsize $\pm$1.12} & 18.24{\scriptsize $\pm$0.36} & 14.40{\scriptsize $\pm$3.11} & 8.17{\scriptsize $\pm$0.78} & 5.17{\scriptsize $\pm$0.57} & 5.58{\scriptsize $\pm$0.81} \\
			CoMatch \cite{li2021comatch} & 35.08{\scriptsize $\pm$0.69} & 25.23{\scriptsize $\pm$0.50} & 15.12{\scriptsize $\pm$1.88} &9.56{\scriptsize $\pm$1.35} & 5.75{\scriptsize $\pm$0.43} & \underline{4.81}{\scriptsize $\pm$1.05} \\
			SimMatch \cite{zheng2022simmatch} & 23.78{\scriptsize $\pm$1.08} & 17.06{\scriptsize $\pm$0.78} & \underline{11.77}{\scriptsize $\pm$3.20} & 7.55{\scriptsize $\pm$1.86} & 7.66{\scriptsize $\pm$0.60} & 5.27{\scriptsize $\pm$0.89} \\
			SoftMatch \cite{chen2023softmatch} & 22.67{\scriptsize $\pm$1.32} & 16.84{\scriptsize $\pm$0.66} & 13.55{\scriptsize $\pm$3.16} & 7.84{\scriptsize $\pm$1.72} & 5.75{\scriptsize $\pm$0.62} & 5.90{\scriptsize $\pm$1.42} \\
			FreeMatch \cite{wang2023freematch} & \underline{21.40}{\scriptsize $\pm$0.30} & \underline{15.65}{\scriptsize $\pm$0.26} & 12.73{\scriptsize $\pm$3.22} & 8.52{\scriptsize $\pm$0.53} & 6.50{\scriptsize $\pm$0.78} & 5.78{\scriptsize $\pm$0.51} \\           
			RegMixMatch & \textbf{19.26}{\scriptsize $\pm$0.56} & \textbf{15.55}{\scriptsize $\pm$0.31} & \textbf{10.11}{\scriptsize $\pm$3.20} & \textbf{7.10}{\scriptsize $\pm$0.66} & \textbf{4.25}{\scriptsize $\pm$0.77} & \textbf{3.61}{\scriptsize $\pm$0.75} \\
			\bottomrule
		\end{tabular}
	}
        \caption{\small Error rates on CIFAR100, STL10 and EuroSAT datasets in USB. The best results are highlighted with \textbf{Bold} and the second-best results are highlighted with \underline{underline}.}
	\label{tab:vit}
\end{table*}

We also validate the performance of RegMixMatch on the USB \cite{wang2022usb} benchmark. USB adopts the pre-trained Vision Transformers (ViT) \cite{dosovitskiy2020image} instead of training ResNets \cite{he2016deep} from scratch for CV tasks, thus achieving better performance with less training time.

We evaluate the performance of RegMixMatch on three datasets, including CIFAR-100 \cite{krizhevsky2009learning}, STL-10 \cite{coates2011analysis}, and Euro-SAT \cite{helber2019eurosat}, under different labeled data conditions. The experimental results are compared against 11 established SSL algorithms, including VAT \cite{miyato2018virtual}, Mean Teacher \cite{tarvainen2017mean}, MixMatch \cite{berthelot2019mixmatch}, ReMixMatch \cite{berthelot2019remixmatch}, UDA \cite{xie2020unsupervised}, FixMatch \cite{sohn2020fixmatch}, FlexMatch \cite{zhang2021flexmatch}, CoMatch \cite{li2021comatch}, SimMatch \cite{zheng2022simmatch}, SoftMatch \cite{chen2023softmatch}, and FreeMatch \cite{wang2023freematch}. For a fair comparison, we use the same setup presented in USB.

Table \ref{tab:vit} shows the results of SSL algorithms on pre-trained backbones, where our RegMixMatch achieves comprehensive superiority. Notably, RegMixMatch still achieves significant improvement on the satellite dataset EuroSAT, where the accuracy of SSL algorithms is close to saturation.

\subsection{ImageNet}

\begin{table}[t]
	\centering
		\begin{tabular}{l|cc|cc}
			\toprule
			Dataset &  \multicolumn{4}{c}{ImageNet} \\
			\cmidrule(r){1-1} \cmidrule(l){2-5}
			\# Label   &  \multicolumn{2}{c|}{1w} & \multicolumn{2}{c}{10w} \\ \cmidrule{1-5}
			 Top-\textit{n} acc   &  \multicolumn{1}{c}{Top-1} & \multicolumn{1}{c|}{Top-5} &  \multicolumn{1}{c}{Top-1} & \multicolumn{1}{c}{Top-5}\\ \cmidrule{1-5}
			FixMatch & 53.61 & 75.96 & 71.53 & 90.36 \\
			FlexMatch & 54.21 & 76.80 & 72.17 & 90.59 \\
			FreeMatch & 54.69 & 77.02 & 72.57 & 90.97 \\
			RegMixMatch & \textbf{58.35} & \textbf{81.10} & \textbf{73.66} & \textbf{91.89} \\
			\bottomrule
		\end{tabular}
	\caption{ImageNet accuracy results.}
	\label{tab:Imagenet}
\end{table}

Following USB, we provide an evaluation on ImageNet of MAE pre-trained ViT-B \cite{he2022masked}, comparing RegMixMatch with FixMatch, Flexmatch, and FreeMatch. We train these algorithms using 10 labels per class and 100 labels per-class, i.e., a total of 10,000 labels and 100,000 labels respectively. We set the batch size for both labeled and unlabeled data to 64, use AdamW with a smaller learning rate of 3e-4 and a weight decay of 0.05. Other algorithmic hyper-parameters stay the same as USB. Our method, RegMixMatch, consistently achieves leading performance across various scenarios.

\begin{table*}[t]
	\centering
	\resizebox{\textwidth}{!}{%
		\begin{tabular}{l|cccc|ccc|ccc|cc}
			\toprule
			Dataset & \multicolumn{4}{c|}{CIFAR10}& \multicolumn{3}{c|}{CIFAR100}& \multicolumn{3}{c|}{SVHN} & \multicolumn{2}{c}{STL10} \\ \cmidrule(r){1-1}\cmidrule(lr){2-5}\cmidrule(lr){6-8}\cmidrule(lr){9-11}\cmidrule(l){12-13}
			
			\# Label & \multicolumn{1}{c}{10} & \multicolumn{1}{c}{40} & \multicolumn{1}{c}{250}  & \multicolumn{1}{c|}{4000} & \multicolumn{1}{c}{400}  & \multicolumn{1}{c}{2500}  & \multicolumn{1}{c|}{10000} & \multicolumn{1}{c}{40}  & \multicolumn{1}{c}{250}   & \multicolumn{1}{c|}{1000} & \multicolumn{1}{c}{40}  & \multicolumn{1}{c}{1000}\\ \cmidrule(r){1-1}\cmidrule(lr){2-5}\cmidrule(lr){6-8}\cmidrule(lr){9-11}\cmidrule(l){12-13}
			
			VAT \cite{miyato2018virtual} & 79.81{\scriptsize $\pm$1.17} & 74.66{\scriptsize $\pm$2.12} & 41.03{\scriptsize $\pm$1.79} & 10.51{\scriptsize $\pm$0.12} & 85.20{\scriptsize $\pm$1.40} & 46.84{\scriptsize $\pm$0.79} & 32.14{\scriptsize $\pm$0.19} & 74.75{\scriptsize $\pm$3.38} & 4.33{\scriptsize $\pm$0.12} & 4.11{\scriptsize $\pm$0.20} & 74.74{\scriptsize $\pm$0.38}  & 37.95{\scriptsize $\pm$1.12} \\
			Mean Teacher \cite{tarvainen2017mean} & 76.37{\scriptsize $\pm$0.44} & 70.09{\scriptsize $\pm$1.60} & 37.46{\scriptsize $\pm$3.30} & 8.10{\scriptsize $\pm$0.21} & 81.11{\scriptsize $\pm$1.44} & 45.17{\scriptsize $\pm$1.06} & 31.75{\scriptsize $\pm$0.23} & 36.09{\scriptsize $\pm$3.98} & 3.45{\scriptsize $\pm$0.03} & 3.27{\scriptsize $\pm$0.05}  & 71.72{\scriptsize $\pm$1.45} & 33.90{\scriptsize $\pm$1.37} \\
			MixMatch \cite{berthelot2019mixmatch} & 65.76{\scriptsize $\pm$7.06} & 36.19{\scriptsize $\pm$6.48} & 13.63{\scriptsize $\pm$0.59} & 6.66{\scriptsize $\pm$0.26} & 67.59{\scriptsize $\pm$0.66} & 39.76{\scriptsize $\pm$0.48} & 27.78{\scriptsize $\pm$0.29} & 30.60{\scriptsize $\pm$8.39} & 4.56{\scriptsize $\pm$0.32} & 3.69{\scriptsize $\pm$0.37}  & 54.93{\scriptsize $\pm$0.96} & 21.70{\scriptsize $\pm$0.68} \\
			ReMixMatch \cite{berthelot2019remixmatch} & 20.77{\scriptsize $\pm$7.48} & 9.88{\scriptsize $\pm$1.03} & 6.30{\scriptsize $\pm$0.05} & 4.84{\scriptsize $\pm$0.01} & 42.75{\scriptsize $\pm$1.05} & 26.03{\scriptsize $\pm$0.35} &  20.02{\scriptsize $\pm$0.27} & 24.04{\scriptsize $\pm$9.13} & 6.36{\scriptsize $\pm$0.22}  & 5.16{\scriptsize $\pm$0.31} & 32.12{\scriptsize $\pm$6.24} & 6.74{\scriptsize $\pm$0.14} \\
			UDA \cite{xie2020unsupervised} & 34.53{\scriptsize $\pm$10.69} & 10.62{\scriptsize $\pm$3.75} & 5.16{\scriptsize $\pm$0.06} & 4.29{\scriptsize $\pm$0.07} & 46.39{\scriptsize $\pm$1.59} & 27.73{\scriptsize $\pm$0.21} & 22.49{\scriptsize $\pm$0.23} & 5.12{\scriptsize $\pm$4.27} & 1.92{\scriptsize $\pm$0.05} & 1.89{\scriptsize $\pm$0.01}  & 37.42{\scriptsize $\pm$8.44} & 6.64{\scriptsize $\pm$0.17} \\
			FixMatch \cite{sohn2020fixmatch} & 24.79{\scriptsize $\pm$7.65} & 7.47{\scriptsize $\pm$0.28} & 4.86{\scriptsize $\pm$0.05} & 4.21{\scriptsize $\pm$0.08} & 46.42{\scriptsize $\pm$0.82} & 28.03{\scriptsize $\pm$0.16} & 22.20{\scriptsize $\pm$0.12} & 3.81{\scriptsize $\pm$1.18} & 2.02{\scriptsize $\pm$0.02}  & 1.96{\scriptsize $\pm$0.03} & 35.97{\scriptsize $\pm$4.14} & 6.25{\scriptsize $\pm$0.33} \\
			Dash \cite{xu2021dash} & 27.28{\scriptsize $\pm$14.09} & 8.93{\scriptsize $\pm$3.11} & 5.16{\scriptsize $\pm$0.23} & 4.36{\scriptsize $\pm$0.11} & 44.82{\scriptsize $\pm$0.96} & 27.15{\scriptsize $\pm$0.22} & 21.88{\scriptsize $\pm$0.07} & 2.19{\scriptsize $\pm$0.18} & 2.04{\scriptsize $\pm$0.02} & 1.97{\scriptsize $\pm$0.01}  & 34.52{\scriptsize $\pm$4.30} & 6.39{\scriptsize $\pm$0.56} \\
			MPL \cite{pham2021meta} & 23.55{\scriptsize $\pm$6.01} & 6.62{\scriptsize $\pm$0.91} & 5.76{\scriptsize $\pm$0.24} & 4.55{\scriptsize $\pm$0.04} & 46.26{\scriptsize $\pm$1.84} & {27.71\scriptsize $\pm$0.19} & 21.74{\scriptsize $\pm$0.09} & 9.33{\scriptsize $\pm$8.02} & 2.29{\scriptsize $\pm$0.04} & 2.28{\scriptsize $\pm$0.02}  & 35.76{\scriptsize $\pm$4.83} & 6.66{\scriptsize $\pm$0.00} \\
			FlexMatch \cite{zhang2021flexmatch} & 13.85{\scriptsize $\pm$12.04} & 4.97{\scriptsize $\pm$0.06} & 4.98{\scriptsize $\pm$0.09} & 4.19{\scriptsize $\pm$0.01} & 39.94{\scriptsize $\pm$1.62} & 26.49{\scriptsize $\pm$0.20} & 21.90{\scriptsize $\pm$0.15} & 8.19{\scriptsize $\pm$3.20} & 6.59{\scriptsize $\pm$2.29} & 6.72{\scriptsize $\pm$0.30} & 29.15{\scriptsize $\pm$4.16} & 5.77{\scriptsize $\pm$0.18} \\
			SoftMatch \cite{chen2023softmatch} & - & 4.91{\scriptsize $\pm$0.12} & 4.82{\scriptsize $\pm$0.09} & 4.04{\scriptsize $\pm$0.02} & \underline{37.10}{\scriptsize $\pm$0.77} & 26.66{\scriptsize $\pm$0.25} & 22.03{\scriptsize $\pm$0.03} & 2.33{\scriptsize $\pm$0.25} & - & 2.01{\scriptsize $\pm$0.01} & 21.42{\scriptsize $\pm$3.48} & 5.73{\scriptsize $\pm$0.24} \\
			FreeMatch \cite{wang2023freematch} & \underline{8.07}{\scriptsize $\pm$4.24} & 4.90{\scriptsize $\pm$0.04} & 4.88{\scriptsize $\pm$0.18} & 4.10{\scriptsize $\pm$0.02} & 37.98{\scriptsize $\pm$0.42} & 26.47{\scriptsize $\pm$0.20} & 21.68{\scriptsize $\pm$0.03} & 1.97{\scriptsize $\pm$0.02} & 1.97{\scriptsize $\pm$0.01} & 1.96{\scriptsize $\pm$0.03} & 15.56{\scriptsize $\pm$0.55} & 5.63{\scriptsize $\pm$0.15} \\
			SequenceMatch \cite{nguyen2024sequencematch} & - & \underline{4.80}{\scriptsize $\pm$0.01} & 4.75{\scriptsize $\pm$0.05} & 4.15{\scriptsize $\pm$0.01} & 37.86{\scriptsize $\pm$1.07} & 25.99{\scriptsize $\pm$0.22} & 20.10{\scriptsize $\pm$0.04} & \underline{1.96}{\scriptsize $\pm$0.23} & 1.89{\scriptsize $\pm$0.31} & 1.79{\scriptsize $\pm$0.02} & \underline{15.45}{\scriptsize $\pm$1.40} & 5.56{\scriptsize $\pm$0.35} \\                
			FlatMatch \cite{huang2023flatmatch} & 15.23{\scriptsize $\pm$8.67} & 5.58{\scriptsize $\pm$2.36} & \underline{4.22}{\scriptsize $\pm$1.14} & \underline{3.61}{\scriptsize $\pm$0.49} & 38.76{\scriptsize $\pm$1.62} & \underline{25.38}{\scriptsize $\pm$0.85} & \textbf{19.01}{\scriptsize $\pm$0.43} & 2.46{\scriptsize $\pm$0.06} & \textbf{1.43}{\scriptsize $\pm$0.05} & \textbf{1.41}{\scriptsize $\pm$0.04} & 16.20{\scriptsize $\pm$4.34} & \underline{4.82}{\scriptsize $\pm$1.21} \\
			RegMixMatch & \textbf{4.35}{\scriptsize $\pm$0.08} & \textbf{4.24}{\scriptsize $\pm$0.02} & \textbf{4.21}{\scriptsize $\pm$0.02} & \textbf{3.38}{\scriptsize $\pm$0.05} & \textbf{35.27}{\scriptsize $\pm$0.60} & \textbf{23.78}{\scriptsize $\pm$0.29} & \underline{19.41}{\scriptsize $\pm$0.10} & \textbf{1.81}{\scriptsize $\pm$0.03} & \underline{1.77}{\scriptsize $\pm$0.01} & \underline{1.79}{\scriptsize $\pm$0.02} & \textbf{11.74}{\scriptsize $\pm$0.57} & \textbf{4.66}{\scriptsize $\pm$0.12} \\
			\midrule
			Fully-Supervised    & \multicolumn{4}{c|}{4.62{\scriptsize $\pm$0.05}} & \multicolumn{3}{c|}{19.30{\scriptsize $\pm$0.09}}  & \multicolumn{3}{c|}{2.13{\scriptsize $\pm$0.01}} & \multicolumn{2}{c}{-}\\
			\bottomrule
		\end{tabular}
	}
        \caption{\small Error rates on CIFAR10/100, SVHN, and STL10 datasets. The fully-supervised results of STL10 are unavailable since we do not have label information for its unlabeled data. The best results are highlighted with \textbf{Bold} and the second-best results are highlighted with \underline{underline}.}
	\label{tab:co}
\end{table*}

\end{document}